\documentclass{article}
\pdfoutput=1
\usepackage[margin=1.2in]{geometry}
\usepackage[usenames,dvipsnames]{xcolor}
\usepackage{microtype}
\usepackage{graphicx}
\usepackage{subcaption}
\usepackage{booktabs} 
\usepackage[authoryear,square]{natbib}
\usepackage{tikz}
\usepackage{nicematrix}
\usepackage{enumitem}
\usepackage{algorithm}
\usepackage{algorithmic}
\usepackage{hyperref}

\usepackage{stfloats} 

\usepackage{hyperref}
\usepackage{framed}
\usepackage[most]{tcolorbox}

\usepackage{amsmath}
\usepackage{amssymb}
\usepackage{mathtools}
\usepackage{wrapfig}
\usepackage{amsthm}
\usepackage{bm}
\DeclareUnicodeCharacter{2500}{-}

\newcommand\Regret{\mathrm{Regret}}

\newcommand\BRegret{\mathrm{BayesRegret}}

\usepackage[textsize=tiny]{todonotes}
\usepackage[capitalize,noabbrev,nameinlink]{cleveref}
\usepackage{authblk}

\author[1]{Aida Afshar$^\dagger$$^,$}
\author[1]{Yuke Zhang$^\dagger$$^,$}
\author[1,2]{Aldo Pacchiano}
\affil[1]{Boston University}
\affil[2]{Broad Institute of MIT and Harvard}

\colorlet{shadecolor}{gray!20}

\tcbset{
  colback=gray!10,
  colframe=black!50,
  width=\columnwidth
}
\hypersetup{
    colorlinks = true,
    linkcolor = RedViolet,
    citecolor=NavyBlue,
    }

\newcommand{\refalgbyname}[2]{\hyperref[#1]{\texttt{\textbf{#2}}}}

\theoremstyle{plain}
\newtheorem{theorem}{Theorem}[section]

\newtheorem{lemma}[theorem]{Lemma}
\newtheorem{corollary}[theorem]{Corollary}
\theoremstyle{definition}
\newtheorem{definition}[theorem]{Definition}

\theoremstyle{remark}

\theoremstyle{example}

\title{Bayesian Online Model Selection}

\begin{document}

\maketitle
\begingroup
  \renewcommand{\thefootnote}{†}
  \stepcounter{footnote}
  \footnotetext{Equal Contribution.}
\endgroup

\begin{abstract}
  Online model selection in Bayesian bandits raises a fundamental exploration challenge: When an environment instance is sampled from a prior distribution, how can we design an adaptive strategy that explores multiple bandit learners and competes with the best one in hindsight? We address this problem by introducing a new Bayesian algorithm for online model selection in stochastic bandits. We prove an oracle-style guarantee of $\widetilde{\mathcal{O}} \left( d^\star M \sqrt{T} + \sqrt{MT} \right)$ on the Bayesian regret, where $M$ is the number of base learners, $d^\star$ is the regret coefficient of the optimal base learner, and $T$ is the time horizon. We also validate our method empirically across a range of stochastic bandit settings, demonstrating performance that is competitive with the best base learner. 
  Additionally, we study the effect of sharing data among base learners and its role in mitigating prior mis-specification. 
\end{abstract}
\section{Introduction}\label{sec:introduction}
The stochastic bandit problem is a fundamental model for interactive decision-making \citep{Lattimore_Szepesvári_2020}. At each round of interaction, a learner selects one of several actions, each associated with an unknown reward distribution, and observes a random reward drawn from that distribution. The performance of the learner is commonly measured in terms of \emph{regret}: the difference between the cumulative reward of the chosen actions and that of the best possible strategy in this environment. The objective is to design algorithms that achieve sublinear regret, ensuring that the learner's averaged regret vanishes as it is allowed to interact more with the environment. This simple yet powerful framework captures the essence of learning from (partial) feedback and lies at the heart of many real-world applications such as recommendation systems, robotics, and finance \citep{silva2022multi, ni2023contextual, klarich2024thompson}. 

In its Bayesian formulation, the stochastic bandit problem posits a prior distribution over the unknown reward distributions, capturing existing beliefs about the environment. This perspective motivates powerful algorithms such as Thompson Sampling \citep{thompson1933likelihood, russo2020tutorialthompsonsampling}, which induces exploration by sampling from the posterior and selecting an action that is optimal for the sampled instance \citep{pmlr-v23-agrawal12}. Bayesian bandits are particularly appealing when prior knowledge can be incorporated and updated via tractable posterior inference, as in the presence of conjugate priors. When conjugacy is unavailable and exact inference is intractable, one typically resorts to approximation: Markov chain Monte Carlo (MCMC) methods, which asymptotically target the true posterior, or deterministic alternatives such as Laplace approximations and variational inference, which provide tractable surrogate posteriors \citep{BO2016, pmlr-v119-mazumdar20a}.

In this paper, we study the online model selection problem originally introduced in frequentist setting \citep{agarwal2017corralling, pacchiano2020model, marinov2021pareto}. In this problem, the learner is given a finite collection of bandit algorithms (``models''), which we will refer to as \emph{base learners}. Each base learner is potentially well-suited to a different environment instance, e.g., one base learner may be tuned for a sparse linear structure while another is optimized for low-dimensional generalized linear rewards. We study this problem in the Bayesian setting, where at the beginning of interaction, an environment is sampled from a known prior distribution, and the learner interacts with this fixed instance over time. In each round, the learner selects a base learner, follows its policy to choose an action, and observes the resulting reward.  Our goal is to design a Bayesian model selection strategy with an \emph{oracle-best} guarantee on Bayesian regret: in expectation over the draw of the environment from the prior, the learner’s regret is competitive with that of an oracle that, knowing the realized environment instance, would commit to the single best base algorithm for that instance.


Existing approaches to Bayesian regret minimization, such as Thompson Sampling (TS) and its variants, exhibit strong empirical performance and enjoy sublinear regret guarantees in stochastic bandits \citep{zhang2022feel}. However, they are not tailored to online model selection. Classic TS is designed for decision sets in which each action has a fixed reward distribution; in model selection, by contrast, the learner chooses among base learners, each of which is itself a bandit algorithm whose behavior and performance evolve as it is run. Identifying the best base learner for the realized environment, therefore, requires deliberate meta-exploration across learners, rather than only action-level exploration within a single learner. Moreover, standard TS explores via posterior sampling followed by greedy play with respect to a sampled model, and does not explicitly exploit additional structural signals (e.g., the informational value of certain actions) that may be crucial for quickly distinguishing between competing learners. 

This limitation becomes pronounced in environments with \emph{information locks}, where some actions are uniformly suboptimal under every environment in the prior yet are highly informative about which actions are optimal \citep{brukhim2025hardness, pacchiano2023experiment}. In such settings, specialized strategies that intentionally query informative actions can dramatically outperform standard TS. These observations motivate a Bayesian approach to online model selection that can combine complementary algorithmic behaviors and automatically adapt to the realized environment instance.  While prior literature has developed oracle-style guarantees for model selection in the frequentist setting \citep{ pmlr-v238-dann24a, pmlr-v139-cutkosky21a}, a Bayesian counterpart with comparable theoretical guarantees has remained unexplored. To this end, we propose a new Bayesian algorithm for online model selection in general stochastic bandits. We summarize our contributions as follows,

\begin{itemize}
    \item \textbf{Bayesian Online Model Selection Algorithm.}
    We introduce a Bayesian algorithm for online model selection in stochastic bandits. Unlike earlier frequentist approaches that require candidate regret bounds for each base learner \citep{pacchiano2020regretboundbalancingelimination}, and in the spirit of recent bound-free frequentist procedures \citep{dann2024data}, our method does not assume known regret bounds and instead uses posterior samples to actively estimate and compare the learners' performance.
    
    \item \textbf{Oracle-best Bayesian-regret Guarantee.}
    We prove that our algorithm achieves an oracle-best bound on the Bayesian regret of
    \(\widetilde{\mathcal{O}}\!\left(d^\star M\sqrt{T} + \sqrt{MT}\right)\),
    where \(M\) is the number of base learners, \(d^\star\) is the regret coefficient of the optimal base learner, and \(T\) is the horizon.
    
    \item \textbf{Generalization of Thompson Sampling.}
    We show that our approach can be viewed as a principled generalization of Thompson Sampling \citep{thompson1933likelihood, daniel2018tutorial}, when the base learners are stationary policies that each select a fixed arm in a \(K\)-armed bandit environment, our algorithm recovers the classical \(\widetilde{\mathcal{O}}(\sqrt{KT})\) Bayesian-regret guarantee.
    \item 
    \textbf{Sharing Data Helps Recovering from Mis-specification.} 
    Through various experiments, we show that sharing data among base learners consistently improves the performance of the meta learner. We further show that when the meta learner has a mis-specified prior, allowing access to at least one well-specified base learner helps the meta learner to recover from mis-specification.
\end{itemize}

\section{Preliminaries}

Bayesian sequential decision-making has been vastly studied across a range of settings, including reinforcement learning, contextual bandits, and more general partially observed environments \citep{fu2022model, ghavamzadeh2015bayesian} 


\paragraph{Bayesian Stochastic Bandits} We focus on the stochastic bandit problem \citep{Lattimore_Szepesvári_2020}, where a learner is sequentially interacting with an environment instance $\nu$ with action space $\mathcal{A}$. At each round of interaction, taking action $a \in \mathcal{A}$ yields a reward $r \sim \nu(a)$ with mean $\mu_\nu(a) \in [0,1]$. In the Bayesian formulation, the environment $\nu$ is drawn once at the beginning of the interaction from a known prior distribution $\mathcal{P}$ over environment instances, and then remains fixed for the duration of an episode of length $T \in \mathbb{N}$.

A (possibly randomized) learner produces at each round $t \in [T]$ a distribution $\pi_t \in \Delta(\mathcal{A})$ over actions, samples an action $a_t \sim \pi_t$, and observes a reward $r_t \sim \nu(a_t)$. Let $\mu^\star_{\nu} \coloneqq \max_{a \in \mathcal{A}} \mu_\nu(a)$ denote the optimal mean reward in an environment $\nu$. The performance of a learner is measured by its \emph{Bayesian regret}, abbreviated as Bayes regret over horizon $T$,
\begin{align*}
\label{equation::bayes_regret}
\BRegret_T
\;=\;
\mathbb{E}\!\left[\sum_{t=1}^T \mu^\star_{\nu} - \mu_\nu(a_t) \mid \nu\right]
\end{align*}
where the expectation is over the draws $\nu \sim \mathcal{P}$, and any internal randomization of the learner. The objective is to design algorithms whose Bayesian regret grows sublinearly in $T$. 

\paragraph{Thompson Sampling} Thompson Sampling (TS) is one of the most celebrated approaches to achieve sublinear regret in Bayesian environments \citep{daniel2018tutorial}. Let $\mathcal{H}_{t} \coloneqq \{(a_s,r_s)\}_{s=1}^{t-1}$ denote the history up to round $t$. TS maintains the posterior distribution $\mathcal{P}_t(\cdot) \coloneqq \mathcal{P}_t(\cdot \mid \mathcal{H}_{t})$ over environments. In the case of multi-armed bandit (MAB) problems with finite $K$ arms, TS samples at round $t$ a candidate environment $\tilde{\nu}_t \sim \mathcal{P}_t$ and plays an arm that is optimal for the sampled instance:
\[
a_t \in \arg\max_{a \in [K]} \mu_{\tilde{\nu}_t}(a).
\]
TS is widely used due to its conceptual simplicity and strong empirical performance, and it admits sublinear Bayesian-regret guarantees of order $\tilde{O}(\sqrt{KT})$
in standard MAB models.
\section{Problem Setup}
\label{sec:bayesianmodsel}
In online model selection, a \textit{meta learner} interacts with a Bayesian bandit environment $\nu\sim\mathcal{P}$ while having access to a finite set of $M$ \textit{base learners}, indexed by $[M]$. Each base learner $i$ maintains an internal state and, when queried at round $t$, outputs a policy $\pi_t^i \in \Delta(\mathcal{A})$. The interaction protocol is as follows. At each round $t$, the meta learner,
\begin{enumerate}
    \item Selects a base learner index $i_t \in [M]$;
    \item Plays action $a_t \sim \pi_t^{i_t}$ and observes reward $r_t \sim \nu({a_t})$;
    \item Updates its own selection rule, and passes $(a_t, r_t)$ to the selected base learner $i_t$ to update its internal state.
\end{enumerate}

We denote $\mathcal{I}_t^{i} =  \{ l \in [t-1] : i_l = i \}$ as the set of timesteps that base learner $i$ has been selected up to $t \in [T]$. We use $n_t^{i} = \sum_{l=1}^{t-1} \mathbf{1}[i_l = i] = |\mathcal{I}_t^i|$ as the number of rounds that base learner $i$ has been selected up to time $t \in [T]$. Finally, we define $ u_t^i = \sum_{l=1}^{t-1} \mathbf{1}[i_l=i]\cdot r_l$
and $\bar{u}_t^i = \sum_{l=1}^{t-1} \mathbf{1}[i_l=i] \cdot\mathbb{E}[r \mid \pi^i_l]$.

In Bayesian online model selection, we evaluate the meta learner's performance by its Bayesian regret. For a fixed environment $\nu \in \text{Supp}(\mathcal{P})$, the regret of the meta learner is equal to the sum of the regret of the base learners, 
\begin{align*}
     \Regret_T(\nu) = \sum_{i=1}^{M} \Regret_T^i(\nu)
\end{align*}
where $\Regret_T^i(\nu)$ is the regret of base learner $i$ after $T$ rounds of interaction with environment $\nu$,
\begin{align*}
    \Regret_T^i(\nu) =  \sum_{t \in \mathcal{I}_T^i} \mu^{\star}_{\nu} - \mu_{\nu} (a_t)
\end{align*}
Finally, we calculate the Bayesian regret of meta learner by taking the expectation over the prior and the internal randomness of the learners, 
\begin{align*}
    \BRegret_T 
     = \mathbb{E} \left[ \sum_{i=1}^{M} \Regret_T^i(\nu) \mid \nu \right]
    = \mathbb{E} \left[ \sum_{i=1}^{M} \sum_{t \in \mathcal{I}_T^i } \mu^{\star}_{\nu} - \mu_{\nu}(a_t) \mid \nu \right] 
\end{align*}

The goal in Bayesian online model selection is to design a selection strategy whose Bayesian regret is comparable to the performance of an oracle that, for each realized environment instance $\nu$, would commit to the single best base learner for that instance. To formalize this objective, we give the following definition,

\begin{definition}[Regret Coefficient]
    \label{def::BRC}
    Let $\nu \sim \mathcal{P}$ be the environment sampled from the Bayesian prior, and let $\delta \in (0, 1)$. For any $t \in [T]$, the regret coefficient of base learner $\mathcal{B}^i, \, i \in [M]$,  is the minimum value $d_t^i(\nu, \delta)$ that satisfies,
    \begin{align*}
        \Regret_l^i(\nu) \leq d_t^i(\nu, \delta) \sqrt{l},  \quad \quad \forall l \leq t
    \end{align*}
    with probability at least $ 1-\delta$. 
\end{definition}

The regret coefficient quantifies how the regret of each base learner scales with $t$. For instance, an ill-performing base learner with linear regret has a regret coefficient that grows at the rate $\sqrt{t}$. 

\begin{tcolorbox}{\textbf{Bayesian Online Model Selection Objective}}
 \label{obj::modsel_guarantee}
 
 We are interested in the following guarantee on the Bayesian regret of the meta learner, 
\begin{align*}
    \BRegret_T \leq \tilde{\mathcal{O}} \left(\text{Poly}(d^\star) \, \sqrt{T} \right)
\end{align*}
where $d^\star$ is defined as,
\begin{align*}   
    \label{def::d_star}
    d^\star := \max_{\nu \sim \mathcal{P}} 
     \min_{i \in [M]} d^i_T(\nu, \delta) 
\end{align*}
\end{tcolorbox}

Online model selection has been a long standing problem in the literature of sequential decision making, and several works has studies this problem in the frequentist setting. Using the candidate regret bounds of base learners, \citep{pacchiano2020regretboundbalancingelimination, pmlr-v139-cutkosky21a} proposed a regret balancing method satisfying $\Regret_T = \tilde{\mathcal{O}}\left(d^\star M\sqrt{T} + {d^\star}^2 \sqrt{MT} \right)$. Later \citep{dann2024data} achieved the same bound without requiring the knowledge of candidate regret bounds, and instead estimated the bounds in a data-driven manner. 

\begin{figure*}[t]
\begin{minipage}{0.95\textwidth}
\begin{algorithm}[H]
    \caption{B-MS: Bayesian Online Model Selection Algorithm}\label{alg:modselTS}
    \begin{algorithmic}[1]
    \REQUIRE Prior $\mathcal{P}$, base learners $ \{ \mathcal{B}^i, i \in [M]\}$, horizon $T$,  $n_1 = [n_1^1,\dots, n_1^M]^\top=0_M$, $\mathcal{I}_1^i=\emptyset$ for $i\in[M]$, and  $\mathcal{P}_1=\mathcal{P}$
    \FOR{$t = 1, 2, ..., T$}
        \IF{$t\leq M$}
            \STATE Select base learner $i_t = t \bmod M$ 
        \ELSE 
            \STATE Sample $\tilde{\mu}_t \sim \mathcal{P}_t(\cdot\mid \mathcal{H}_{t})$   
            \STATE Set $\tilde{\mu}_t^\star = \max_{a \in \mathcal{A}} \tilde{\mu}_t(a)$
            \STATE Calculate $\phi_t(i) =  n_{t}^i \tilde{\mu}_t^\star - \sum_{l \in \mathcal{I}_{t}^i} \tilde{\mu}_t(a_l)$ 
            \STATE Select base learner $i_t = \arg\min_{i \in [M]} \phi_t(i)$
        \ENDIF
        \STATE Play action $a_t \sim \pi^{i_t}$ and observe reward $r_t$ 
        \STATE Pass ($a_t, r_t$) to $\mathcal{B}^{i_t}$ to update its policy
        \STATE Update statistics $$n^{i_t}_{t+1} \leftarrow n^{i_t}_{t} + 1,\quad  n^{i'}_{t+1} \leftarrow n^{i'}_{t} \text{ for } i'\neq i_t, \quad \text{ and } \quad  \mathcal{I}_{t+1}^i=  \{ l \in [t] : i_l = i \} \text{ for } i\in[M]$$
        \STATE Update history $\mathcal{H}_{t+1}\leftarrow\mathcal{H}_{t}\cup \{a_t, r_t\}$ and  posterior distribution $\mathcal{P}_{t+1}(\cdot\mid \mathcal{H}_{t+1})$
    \ENDFOR
    \end{algorithmic}
\end{algorithm}
\end{minipage}
\end{figure*}

In the following section, we introduce the Bayesian online model selection algorithm and prove that it satisfies $\BRegret_T = \tilde{\mathcal{O}}(d^\star M\sqrt{T}+ \sqrt{MT} )$.  To the best of our knowledge, this is the first Bayesian method for online model selection with oracle-best guarantees. Similar to \citep{dann2024data}, our method does not require knowing the candidate regret bounds. It leverages the posterior distribution to estimate the performance of base learners. 

\section{Methodology \& Algorithm}
\label{sec::algorithm}

The Bayesian Online Model Selection algorithm uses the posterior samples to estimate the performance of base learners throughout training. At time $t \in [T]$, the meta learner samples a mean reward corresponding to each action from the posterior distribution, 
\begin{align*}
    \tilde{\mu}_t(a) &\sim \mathcal{P}_t(\cdot \mid \mathcal{H}_{t}), \quad\quad \forall a \in \mathcal{A}
\end{align*}
with the optimal sampled mean reward denoted as $\tilde{\mu}_t^\star = \arg \max_{a \in \mathcal{A}} \tilde{\mu}_t (a)$.\\
The \textit{balancing potential} of base learner $\mathcal{B}^i \text{ for } i \in [M]$, at time $t \in [T]$, is defined as follows,
\begin{align*}
    \mathcal{\phi}_t(i) =\sum_{l \in \mathcal{I}_t^i} \tilde{\mu}_t^\star - \tilde{\mu}_t(a_l) =  n_t^i \tilde{\mu}_t^\star - \sum_{l \in \mathcal{I}_t^i} \tilde{\mu}_t(a_l).
\end{align*}
This choice of potential function has several properties. The function $\phi_t(i)$ mirrors the definition of the regret of base learner $\mathcal{B}^i$ at time $t$. In fact, if we calculate the potential using the true mean rewards instead of sampled means, we get the realized regret of base learner $\mathcal{B}^i$. Since we do not know the true mean rewards, we leverage the sampled mean rewards drawn from the posterior. As the posterior receives more samples and gets concentrated around the true reward distributions, the potential gives better and better estimates of the true regret of each base learner. 

The algorithm then selects $i_t$ to be the base learner with minimum balancing potential at time $t$, 
\begin{align*}
    i_t & = \arg\min_{i \in [M]} \phi_t(i) 
\end{align*}
The meta learner selects the base learner with the minimum estimated regret, exploiting the current best candidate. At the same time, if a base learner has not been explored enough, it will have lower estimated regret and will be selected by the meta learner under this selection rule.  

Finally, after selecting the base learner, the meta learner plays the action suggested by the policy of this base learner $a_t \sim \pi^{i_t}$, and observes reward $r_t$. The action and reward will be passed to the base learner to update its policy, and will also be used to update the posterior distribution.

A key property of the Bayesian Online Model Selection algorithm is that it keeps a global posterior distribution across the base learners, using the action-reward pairs gathered by all of the base learners to update the posterior. As a result, even though the base learners do not communicate, they share information via the posterior distribution, gaining statistical efficiency.  The pseudo-code of the Bayesian Online Model Selection algorithm is depicted in Algorithm \ref{alg:modselTS}.

\section{Analysis}
\label{sec::analysis}
In the following sections, we provide the theoretical analysis of Algorithm \ref{alg:modselTS}. We first state our assumptions, and some of the key lemmas that we use in our analysis. We then provide a proof sketch of the oracle-best guarantee and refer the reader to Appendix \ref{sec::appendix_proof} for the full proof.

\subsection{Assumptions}
\label{subsec::assumption}
We consider the following assumptions in our analysis: 
\begin{enumerate}
    \item \textbf{(Boundedness)} We assume the rewards are bounded $r \in [0, 1]$ at all timesteps.
    \item \textbf{(Well-Specified Prior)} We assume that the meta learner has access to the correct prior distribution $\mathcal{P}$.
\end{enumerate}


\subsection{Key Lemmas}
 
\begin{lemma}[Good Event]
    \label{lemma::good_event}
    Let $\delta \in (0, 1)$. Define the event, 
    \begin{align*}
        \mathcal{E}_{\text{good}} = \left\{ \left| u_t^i - \overline{u}_t^i \right| \leq c \sqrt{n_t^i \log\left(\frac{tM}{\delta}\right)} \quad \forall i \in [M] \right\}
    \end{align*}
    Then, there is an absolute constant $c$ such that the event $\mathcal{E}_{good}$ holds with probability at least $ 1 - \delta$.
    \proof Appendix \ref{lemma::good_event_proof}

\end{lemma}

\begin{lemma}
\label{lemma::key_lemma}
Denote $i_t^\star = \arg\min_{i \in [M]} \Regret_t^i$ as the base learner with minimum regret at time $t \in [T]$, and let $\mathcal{F}_{t}$ be the $\sigma$-algebra induced by all variables up to round $t$. For any function $f_t(i): [M] \rightarrow \mathbb{R} $ that is $\mathcal{F}_{t}$-measurable,
\begin{align*}
    \mathbb{E} \left[ f_t(i_t^\star) \mid \mathcal{F}_{t} \right] = \mathbb{E} \left[ f_t(i_t) \mid \mathcal{F}_{t} \right]
\end{align*}
\proof Appendix \ref{lemma::key_lemma_proof}
\end{lemma}

\begin{lemma}
    \label{lemma::UCB}
    Suppose the event $\mathcal{E}_{good}$ holds, and denote, 
    \begin{align*}
        f_t(i) = \frac{u_t^i}{n_t^i} +  c \sqrt{\frac{\log(tM/\delta)}{n_t^i}} + \frac{d^\star}{\sqrt{n_t^i}}
    \end{align*}
    For an optimal base learner $i_t^\star$, the function $f_t(i_t^\star)$ overestimate the true mean $\mu^\star$,
    \begin{align*}
        \mu^\star \leq f_t(i_t^\star)
    \end{align*}
    for all $t \in [T]$.
    \proof Appendix \ref{lemma::UCB_proof}
\end{lemma}

\begin{lemma}
\label{lemma::helper_lemma}
Under event $\mathcal{E}_{good}$, the following holds,
\begin{align*}
         & \mathbb{E}_{\nu} \left[  \sum_{i=1}^M  \sum_{t \in \mathcal{I}_T^i} \frac{u_t^i}{n_t^i} - \mu_t^i  \right]
         \leq d^\star M \sqrt{T}  +  3c \sqrt{MT \log \left(\frac{MT}{\delta} \right)} + 2M^2T^2 \delta
\end{align*}
\proof Appendix \ref{theorem::Bayesian_Regret_Bound_proof}
\end{lemma}

\begin{theorem}[Oracle-Best Guarantee]
\label{theorem::Bayesian_Regret_Bound}

The Bayesian Regret of Algorithm \ref{alg:modselTS} satisfies, 
\begin{align*}
    \BRegret_T \leq  \widetilde{\mathcal{O}}\left(  d^\star M \sqrt{T} + \sqrt{MT} \right) 
\end{align*}
\proof We provide a proof sketch here, and refer the reader to  Appendix \ref{theorem::Bayesian_Regret_Bound_proof} for the full proof. For clarity of writing, we drop the dependence on environment $\nu$ wherever it is clear from the context and abbreviate the conditioning  as $\mathbb{E}[\cdot \mid \nu] = \mathbb{E}_{\nu} [\cdot]$. 

Recall the definition of the Bayesian regret of the meta learner,  
\begin{align*}
    \BRegret_T  =\mathbb{E} \left[ \sum_{i=1}^{M} \sum_{t \in \mathcal{I}_T^i } \mu^{\star}_{\nu} - \mu_{\nu}(a_t) \mid \nu \right]= \mathbb{E}_{\nu} \left[ \sum_{i=1}^M \sum_{t \in \mathcal{I}_T^i} \mu^\star - \mu^i_t \right]  
\end{align*}

We start by adding and subtracting the function $f_t(i)$ defined in Lemma \ref{lemma::UCB}.
\begin{align*}
    = \mathbb{E}_{\nu} \left[ \sum_{i=1}^M \sum_{t \in \mathcal{I}_T^i} \mu^\star - f_t(i) + f_t(i) - \mu^i_t\right]
\end{align*}

We decompose this regret into two terms and bound each term separately.
\begin{align*}
    = \underbrace{ \mathbb{E}_{\nu} \left[  \sum_{i=1}^M \sum_{t \in \mathcal{I}_T^i}  \mu^\star - f_t(i) \right] }_{\text{(I)}} + \underbrace{  \mathbb{E}_{\nu} \left[  \sum_{i=1}^M \sum_{t \in \mathcal{I}_T^i} f_t(i) - \mu_t^i \right] }_{\text{(II)}}
\end{align*}

The term $\mathrm{(I)}$ can be bounded by first conditioning on filtration $\mathcal{F}_t$, and then using the property that we proved in Lemma \ref{lemma::key_lemma}. By the law of total expectation, 
\begin{align*}
    \mathbb{E}_{\nu} \left[  \sum_{i=1}^M \sum_{t \in \mathcal{I}_T^i}  \mu^\star - f_t(i) \right] 
    = \mathbb{E}_{\nu} \left[  \sum_{i=1}^M \sum_{t \in \mathcal{I}_T^i} \mathbb{E} \left[ \mu^\star - f_t(i) \mid \mathcal{F}_{t} \right] \right] 
\end{align*}
By Lemma \ref{lemma::key_lemma}, $\mathbb{E} \left[ f_t(i_t) \mid \mathcal{F}_{t} \right] = \mathbb{E} \left[ f_t(i_t^\star) \mid \mathcal{F}_{t} \right] $, 
\begin{align*}
    = \mathbb{E}_{\nu} \left[  \sum_{i=1}^M \sum_{t \in \mathcal{I}_T^i} \mathbb{E} \left[ \mu^\star - f_t(i_t^\star)  \mid \mathcal{F}_{t} \right] \right] \leq 0
\end{align*}
The inequality holds due to Lemma \ref{lemma::UCB}, where we showed $\mu^\star \leq f_t(i_t^\star)$ for all $t \in [T]$. It remains to bound the term $\mathrm{(II)}$. Substitute the definition of $f_t(i)$,
\begin{align*}
    &\mathbb{E}_{\nu} \left[  \sum_{i=1}^M \sum_{t \in \mathcal{I}_T^i} f_t(i) - \mu_t^i \right] = \mathbb{E}_{\nu}  \left[  \sum_{i=1}^M \sum_{t \in \mathcal{I}_T^i} \frac{u_t^i}{n_t^i} +  c \sqrt{\frac{\log(tM/\delta)}{n_t^i}} + \frac{d^\star}{\sqrt{n_t^i}} - \mu_t^i \right]
\end{align*}
Rearrange and decompose, 
\begin{align*}
    & \leq \mathbb{E}_{\nu} \left[  \sum_{i=1}^M  \sum_{t \in \mathcal{I}_T^i} \frac{u_t^i}{n_t^i} - \mu_t^i  \right] + \mathbb{E}_{\nu}  \left[  \sum_{i=1}^M \sum_{t \in \mathcal{I}_T^i} c \sqrt{\frac{\log(MT/\delta)}{n_t^i}} + \frac{d^\star}{\sqrt{n_t^i}}  \right]
\end{align*}
The first term is bounded by Lemma \ref{lemma::helper_lemma}. The second term can be bounded by first bounding the sum $\sum_{l = 1}^{n_T^i} \frac{1}{\sqrt{l}} \leq \mathcal{O}\left(\sqrt{n_T^i}\right)$, and then applying the Cauchy-Schwarz inequality. Combining both bounds, we get 
\begin{align*}
    & \leq 2d^\star M \sqrt{T}  +  4c \sqrt{MT \log \left(\frac{MT}{\delta} \right)} + 2M^2T^2 \delta
\end{align*}
Setting $\delta = \tfrac{1}{M^2T^2}$, we get the desired bound.
\end{theorem}

\subsection{Understanding the Theoretical Guarantee}
\label{subsec::guarantee_discription}
The model selection guarantee of $$\BRegret (T) \leq \widetilde{\mathcal{O}}\left( d^\star M\sqrt{T} + \sqrt{MT}\right)$$ that we proved in Theorem \ref{theorem::Bayesian_Regret_Bound} have the following scaling laws with respect to each variable:
\begin{itemize}
    \item Horizon $T$: The bound grows sub-linearly, with a rate of $\sqrt{T}$. This dependence is optimal as it matches the $\sqrt{T}$ barrier of online model selection \citep{pacchiano2020model}. 
    
    \item Optimal Regret Coefficient $d^\star$: The bound scales linearly in $d^\star$, satisfying the target rate of $\mathcal{O}\left( \text{Poly}(d^\star) \sqrt{T}\right)$. Prior work of \citep{marinov2021pareto} has established a lower bound of $\Omega(d{^{\star}}^2 \sqrt{T})$ for online model selection in the frequentist setting. The improved linear dependence on $d^\star$ in our algorithm confirms that Bayesian algorithms can exploit prior knowledge to do better on average than what frequentist bounds would suggest. In comparison to the bounds of other online model selection algorithms, our bound is better than the optimal $\mathcal{O}\left( {d^\star}^2 \sqrt{T}\right)$ rate of the frequentist counterparts, however, one can not directly compare these bounds as one is on Bayesian regret, and the other is on the frequentist regret.


    \item Number of Base Learners $M$: The bound scales linear in the number of base learners $M$. Even though the linear dependence on $M$ matches the prior literature on online model selection \citep{pacchiano2020regretboundbalancingelimination, pmlr-v139-cutkosky21a} as well as the state-of-the-art online model selection algorithms in the frequentist setup \citep{dann2024data}, we conjecture that this rate could be improved to $\sqrt{M}$ using the insights from our current analysis.
\end{itemize}
\section{Comparison with Thompson Sampling}\label{sec:connection2ts}
In this section, we compare the Bayesian Online Model Selection Algorithm \ref{alg:modselTS} with the Thompson Sampling algorithm. Consider the following setup: Suppose we have a $K$-armed bandit environment with mean rewards $[\mu_1 \cdots, \mu_K]$. Consider a base learner that only pulls a fixed arm for all timesteps $t \in [T]$. Then, running Algorithm \ref{alg:modselTS} with base learners $\{\mathcal{B}^1, \cdots, \mathcal{B}^K\}$, where $\mathcal{B}^i$ only pulls arm $i \in [K]$, is equivalent to running Thompson Sampling on this $K$-armed bandit environment. For this setup, the following theorem shows that the Bayesian Online Model Selection algorithm recovers the Bayesian regret bound of Thompson Sampling for multi-armed bandits \citep{russo2020tutorialthompsonsampling, Lattimore_Szepesvári_2020}.

\begin{theorem}
\label{theorem::recovering_TS}

Suppose we have a $K$-armed bandit environment. Consider Algorithm \ref{alg:modselTS} with $K$ base learners, where base learner $i$ pulls fixed arm $i \in [K]$ at all timesteps $t \in [T]$. Then, Bayesian regret of Algorithm \ref{alg:modselTS} recovers the Bayesian regret bound of Thompson sampling by achieving, 
\begin{align*}
    \BRegret_T \leq  \widetilde{\mathcal{O}}\left( \sqrt{KT} \right) 
\end{align*}
\proof Appendix \ref{theorem::recovering_TS_proof}
\end{theorem}
Figure \ref{fig:exp3_averaged_cum_regret_H1000_R1000_v5} in Appendix shows that our algorithm (abbreviated as B-MS), when equipped with a $K$ base learners with different fixed arm selections over time, achieves Bayesian regret that closely overlaps with Thompson Sampling. This provides strong evidence that our Bayesian regret bound coincides with the established bound for TS in this special case $d^\star = 0$, since the optimal base learner always selects the optimal arm resulting zero regret.

\section{Experiments}\label{sec:experiment}
To assess the performance of our algorithm, we conduct experimental studies comparing our proposed meta learner with individual runs of base learners. Given the environment sampled from prior, this framework allows base learners to be different bandit algorithms or the same algorithm with different configurations.

We consider finite $K$-armed bandit setting and sample each bandit environment instance for $R$ times. In model selection framework, we first choose base learners by the meta learner's policy and then choose actions by the base learner's policy.  We assume that a selected base learner $i_t\in[M]$ only select an action at timestep $t$ and pass to next timestep. 

For each experiment, we evaluate: 1) empirical Bayesian regret and 2) optimal action selection rate, reporting both metrics over time with 95\% confidence intervals across runs. 
The former one serves as the classical performance measure, estimating Bayesian regret, where sublinear growth indicates effective learning. 
\[
\widehat{\BRegret}_T
= \frac{1}{R}\sum_{j=1}^R 
\left[\sum_{i=1}^M \sum_{l\in\mathcal{I}_T^i} \mu_{\nu_j}^\star - \mu_{\nu_j}(a_l)\right]
\]
The latter one measures the frequency with which the algorithm identifies and selects the truly best action over different sampled environments, verifying that both meta learner and base algorithms are successfully converging to the optimal choice.
\[
\widehat{\mathrm{OptimalActionSelectRate}}_t
= \frac{1}{R} \sum_{j=1}^{R}
  \mathbf{1}[a_t(\nu_j) = a^\star(\nu_j)]
\]
\subsection{Experiment Setup}
\paragraph{Upper Confidence Bound as Base Learner} 
We consider a multi-armed bandit example, where the learner selects an action and receives a reward from a normal distribution with fixed but unknown mean $\mu^\star\in\mathbb{R}^K$. At each round $t$, 
\(
r_t \sim\mathcal{N}(\mu^\star(a_t), 1).
\)
In Bayesian setting, our reward means are sampled from a given prior, 
\(
\mu^\star\sim\mathcal{N}(0_K, I_{K\times K}),
\)
where $\mu^\star(a)$ is the true mean for action $a$. The policy of each base learner is to select an action that maximizes its Upper Confidence Bound (UCB) value. At timestep $t$, the base learner $i_t$ is selected, its UCB value is calculated, for all $a\in [K]$,
\[
\text{UCB}_t^{i_t}(a) = \hat{\mu}_{t}^{i_t}(a) + c\sqrt{\frac{\log{\left(2K\cdot\sum_{a\in\mathcal{A}}n_{t}^{i_t}(a)/\delta\right)}}{n_{t}^{i_t}(a)}} 
\]
where
\(n_t^i(a) = \sum_{s=1}^{t-1}\mathbf{1}[a_s=a, i_s=i]\),
\(\hat{\mu}_t^i(a) = \sum_{s=1}^{t-1}\bm{1}[a_s=a, i_s=i]\cdot r_s\), and $c$ is confidence radius constant.
We specify different base learners with different values of $c$ selecting from the set $\{0.01, 0.1, 1, 2, 5, 10\}$. The setup is inspired by \citep{pmlr-v238-dann24a}. 
\paragraph{Thompson Sampling as Base Learner}
We consider a two-armed bandit problem. The reward is from Gaussian distribution $\mathcal{N}(\mu^\star(a), 1)$ for $a\in[2]$ where $\mu^\star\sim \mathcal{N}(\mu_0, \sigma_0^2I_{2\times 2})$. The prior mean is $\mu_0 = [0,0.1]$ and the prior standard deviation is $\sigma_0 = 0.05$. We assume prior standard deviations for base learners and meta learners are well-specified.
\begin{figure}[H]
  \centering

  \begin{subfigure}[t]{0.45\linewidth}
    \centering
    \includegraphics[width=\linewidth]{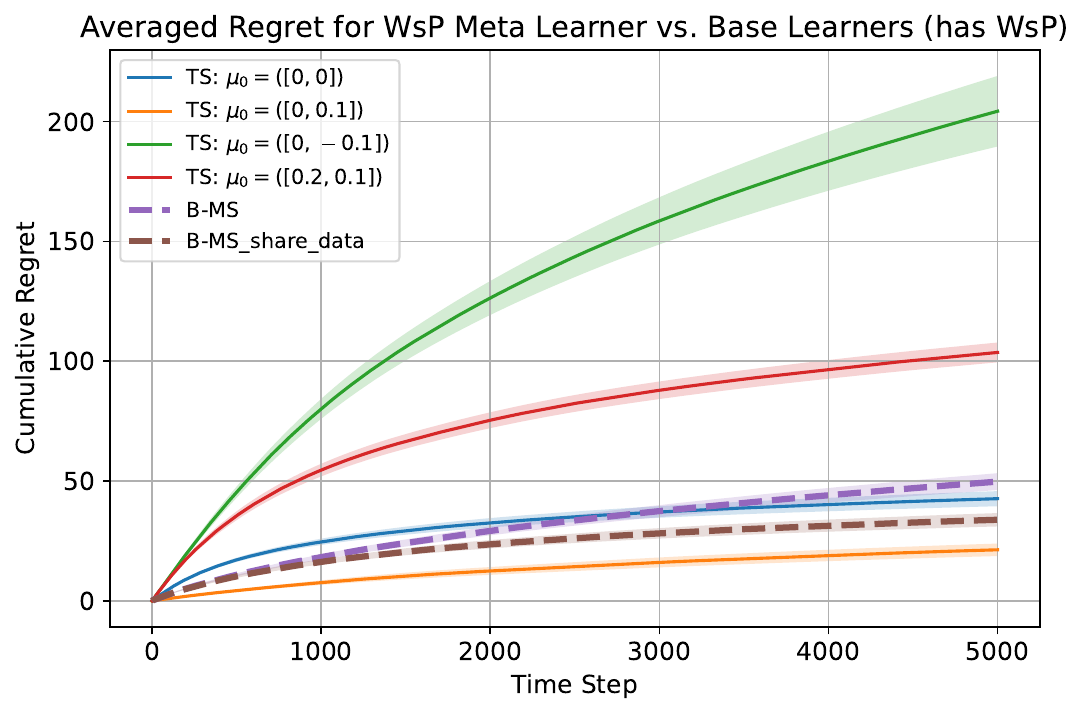}
    \caption{Well-specified meta learner, one well-specified base learner}
    \label{fig:exp14}
  \end{subfigure}
  \hfill
  \begin{subfigure}[t]{0.45\linewidth}
    \centering
    \includegraphics[width=\linewidth]{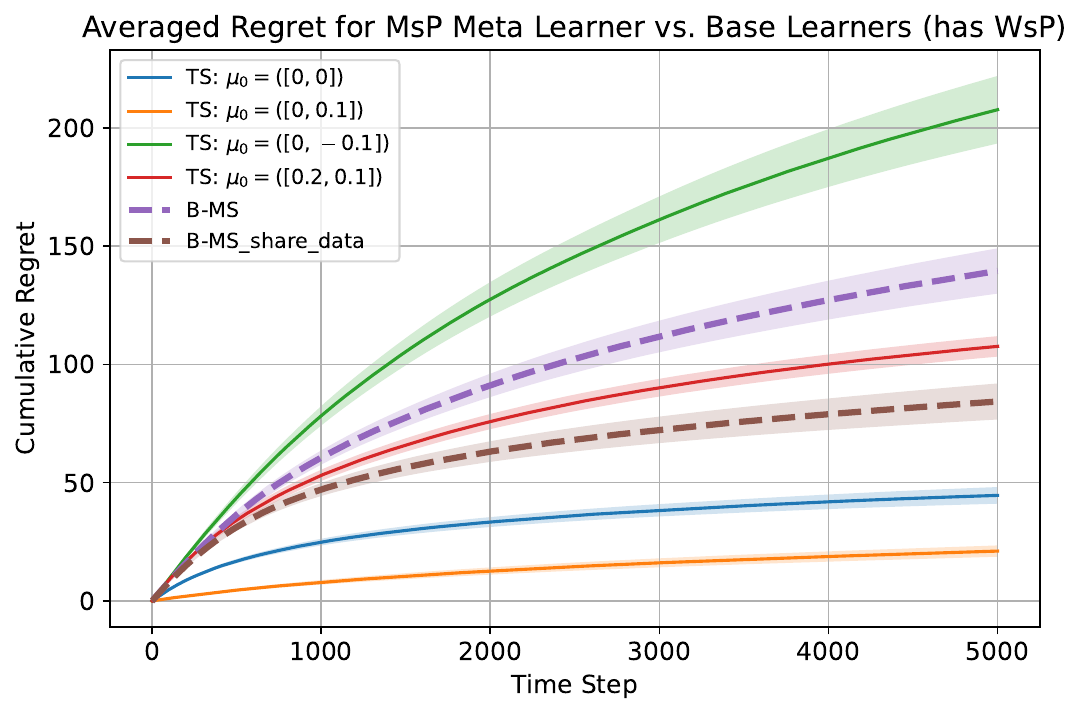}
    \caption{Mis-specified meta learner, one well-specified base learner}
    \label{fig:exp15}
  \end{subfigure}
  \begin{subfigure}[t]{0.45\linewidth}
    \centering
    \includegraphics[width=\linewidth]{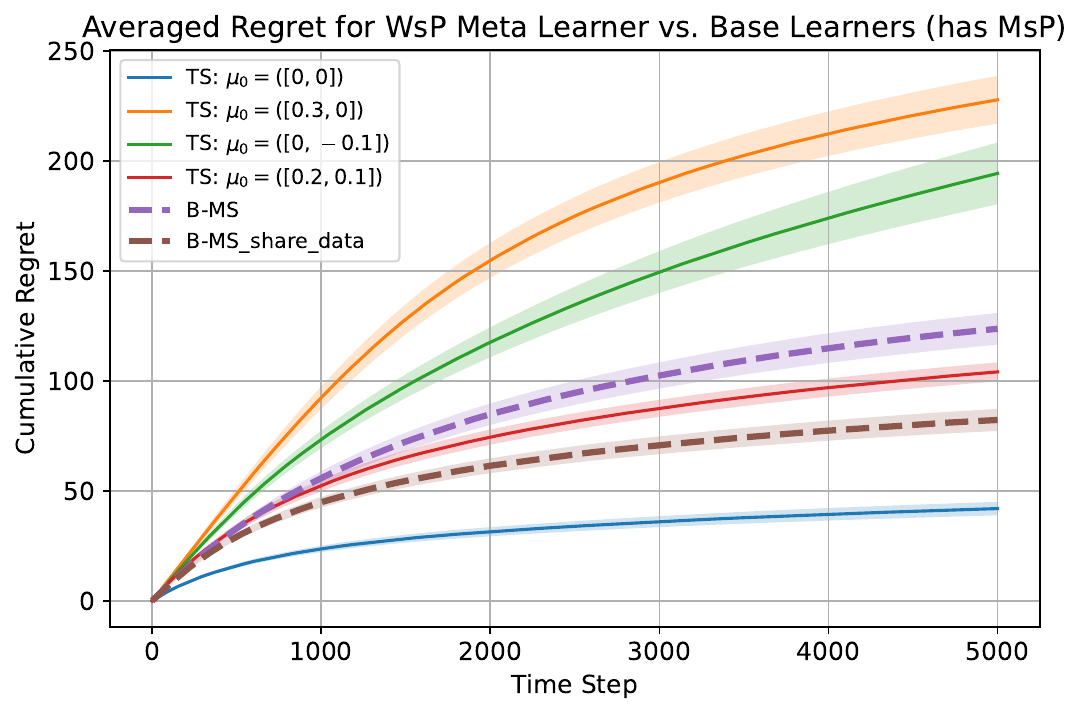}
    \caption{Well-specified meta learner, mis-specified base learners}
    \label{fig:exp16}
  \end{subfigure}
  \hfill
  \begin{subfigure}[t]{0.45\linewidth}
    \centering
    \includegraphics[width=\linewidth]{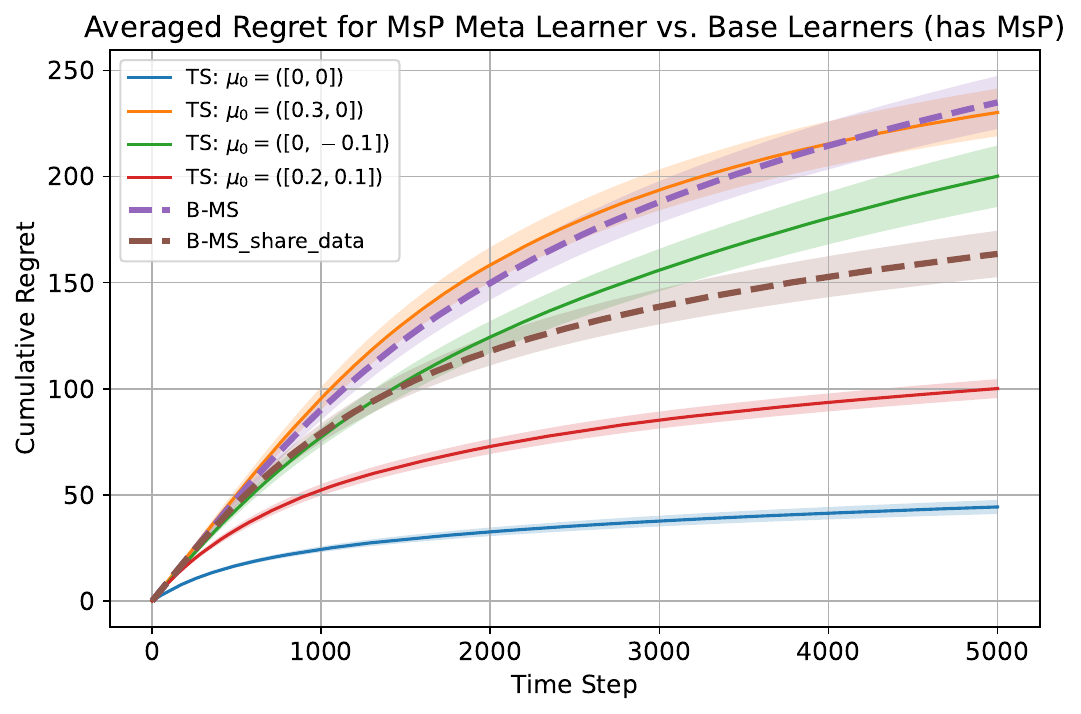}
    \caption{Mis-specified meta learner and base learners}
    \label{fig:exp17}
  \end{subfigure}
  \hfill
  \caption{Empirical Bayesian regret for B-MS vs. TS base learners with different prior specifications ($T=5\times10^3$, $R=500$, $K=2$, $M=4$).}
  \label{fig:exp_all_share_vs_no}
\end{figure}
We set a small prior variance by design, which allowed us to create model mis-specification simply by changing the prior mean. The prior means for TS base learners are chosen from $$\{[0,0], \, [0, 0.1], [0, -0.1], \, [0.2, 0.1], \, [0.3, 0]\}$$ respectively. 
Note that if one of the base learners has prior mean $[0, 0.1]$ (orange line in Figure \ref{fig:exp14}, \ref{fig:exp15}), it is the well-specified base learner. As all base learners has mis-specified prior, we replace it with prior mean $[0.3, 0]$ (orange line in Figure \ref{fig:exp16},  \ref{fig:exp17}). When we specify meta learners to be mis-specified, we choose the prior mean to be $[0, -0.1]$ (base learner with this prior are greenlines in all subfigures in Figure \ref{fig:exp_all_share_vs_no}) which has wrong optimal arm belief at the beginning.

In addition, we compared our original algorithm with a relaxed setting, where data collected by one base learner are shared among all base learners for updating their statistics. In other words, new data ($a_t, r_t$) is used to update $\mathcal{B}^{i}$'s policy for $i\in[M]$ instead of only $\mathcal{B}^{i_t}$'s at each timestep $t$. We refer this setting as data sharing.
\paragraph{Linear Thompson Sampling as Base Learner}
In our linear bandit example, at each time step $t$, the learner selects an $a_t \in \text{UnitBall}(\mathbb{R}^d)$ and receives a reward from a normal distribution
\[
r_t\sim\mathcal{N}(a_t^\top\theta^\star, 1)
\]
And our true parameter $\theta^\star$ is fixed but unknown to the learner and it is sampled from a given prior
\[
\theta^\star\sim\mathcal{N}(0_K, \sigma_0^2I_{K\times K})
\]
We pick $ \sigma_0 = \sqrt{\frac{1}{\lambda}}$ where $\lambda>0$ is regularization parameter. Also $\mu^\star(a) = a^\top\theta^\star$ recovers the true mean given action $a$.
The algorithm
Linear Thompson Sampling (LinTS)~\citep{pmlr-v54-abeille17a} adapts a specific way of selecting arms. 
Once base learner $i_t$ is selected at round $t$, we select 
\[
a_t = \arg\max_{|a|\leq 1} a^\top \tilde{\theta}_t^{i_t}
\]
using sample,
\[\tilde{\theta}_t^{i_t} \sim \mathcal{N}(\hat{\theta}_t^{i_t}, c_i^2d V_t^{i_t,-1})\]
where inverse of covariance matrix $V_t^{i_t}$ and Regularized Least Square (RLS) estimator $\hat{\theta}_t^{i_t}$ are
\begin{align*}
    V_t^{i_t} = \lambda I_{d\times d} + \sum_{l=1}^{\mathcal{I}_t^i} a_{l}a_{l}^\top
    \text{ and }
    \hat{\theta}_t^{i_t} = V_t^{i_t,-1}\left(\sum_{l=1}^{\mathcal{I}_t^i} a_l r_l\right)
\end{align*}
Set dimension $d=10$. We choose hyper-parameter to be confidence radii that are different among base learners $c_i\in\{0, 0.16, 2.5, 5, 25\}$.

\paragraph{Information Lock Example}
We consider a synthetic environment referred to as the information lock, where some suboptimal arms can provide useful information about the optimal arm, analogous to the cheating code example introduced in \citep{foster2023foundations}.

Consider a $(N+K)$-armed bandit environment, where $\mu_1, ..., \mu_N,  \mu_{N+1}, \ldots, \mu_{N+K}$ denote mean rewards of the arms and \[
j^\star = \arg\max_{j \in \{1, \ldots, N+K\}} \mu_j
\] is the optimal arm. Note that the number of regular arms $K$ is chosen to be a power of two and the number of magic arms $N = \log_2 K$. Specifically,
\begin{align*}
\mu_j =
\begin{cases}
    -b_j(j^\star), & j \in \{1, \dots, N\},\\[3pt]
    \frac{1}{2} + \frac{1}{2}\cdot\,\mathbf{1}\{j = j^\star\}, 
    & j \in \{N+1, \dots, N+K\}.
\end{cases}
\end{align*}
Here, the magic arms encode binary representation of the index of best arm,
\begin{align*}
& b(j^\star) = \left(b_1(j^\star), \dots, b_N(j^\star) \right), \\
& \text{where } b_n(j^\star)\in \{0,1\} \quad \forall n\in [N].
\end{align*}
If a learner is aware that the bandit environment follows this structure, it can exploit this information to identify the optimal arm and then select it for the rest of selection. We call this learner an information lock solver (ILS).

\subsection{Result}
Our meta learner achieves performance comparable to the best-performing base learner in both UCB and LinTS bandit settings. The averaged cumulative regret curves track closely with the optimal base learner, while the optimal action selection rates provide compelling evidence: the B-MS algorithm achieves near-identical performance to the best base learner in the UCB setting ($c=1$) and closely approaches the best performer in the LinTS setting ($c=0.16$).
\begin{figure}[H]
  \centering
  \begin{subfigure}[t]{0.45\linewidth}
    \centering
    \includegraphics[width=\linewidth]{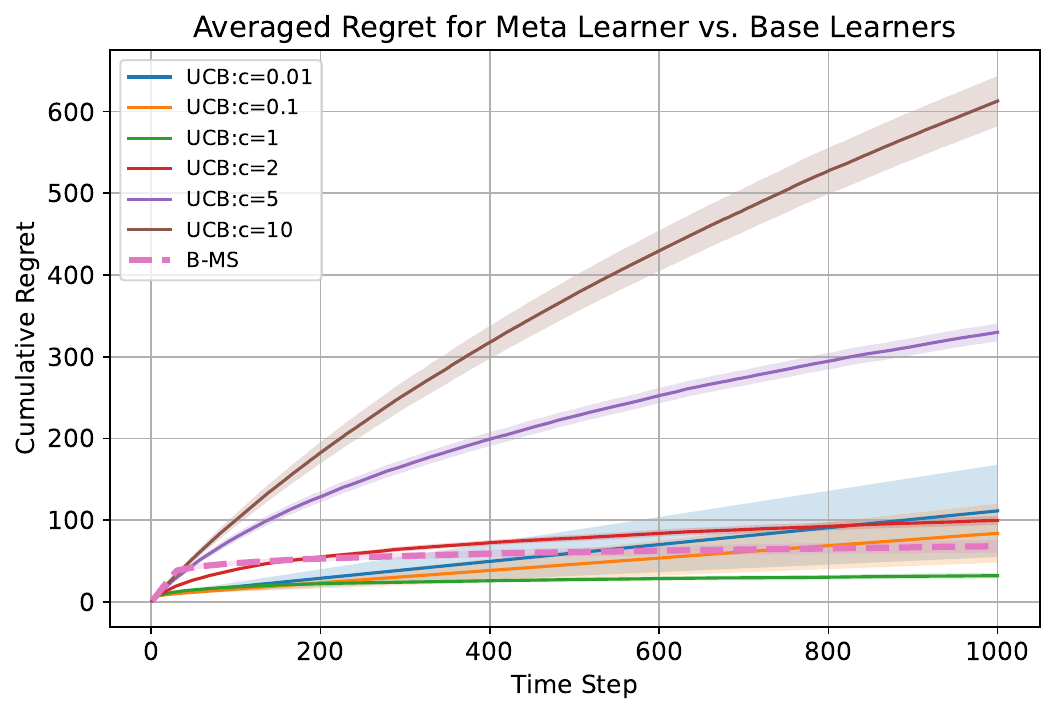}
    \caption{Empirical Bayesian regret}
    \label{fig:exp2-2_averaged_cum_regret_H1000_R100_v5}
  \end{subfigure}
  \hfill
  \begin{subfigure}[t]{0.5\linewidth}
    \centering
    \includegraphics[width=\linewidth]{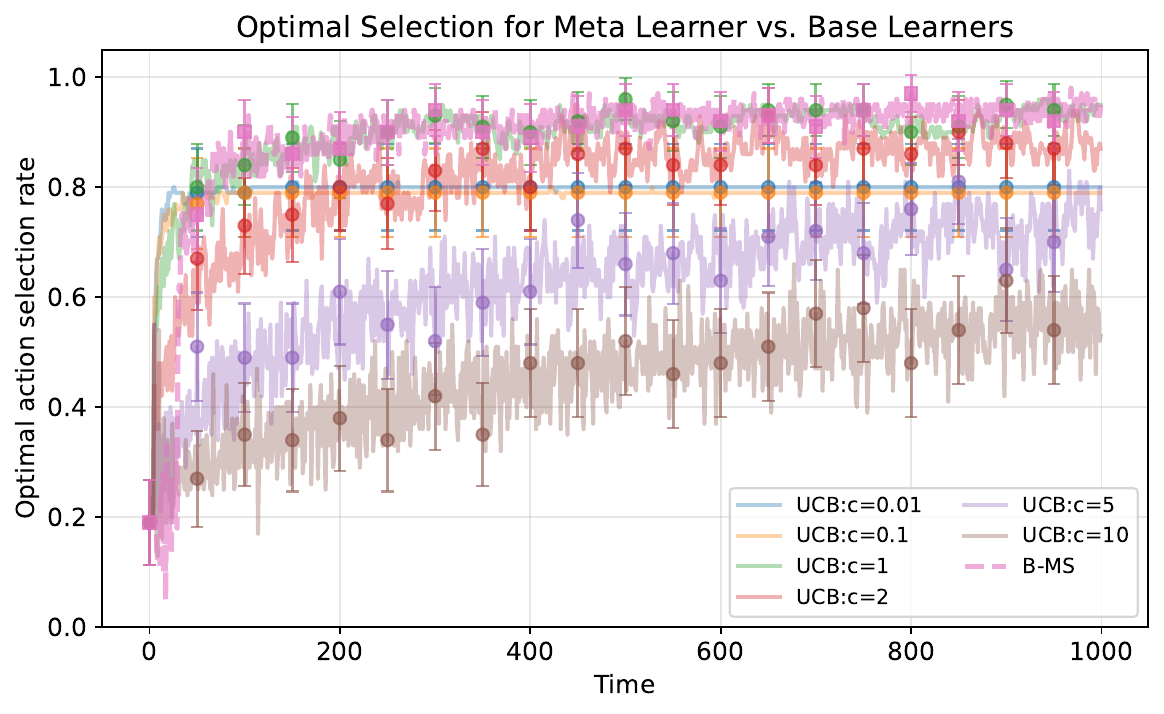}
    \caption{Optimal action selection rate}
    \label{fig:exp2-2_meta_statistics_opt_rate_H1000_R100_v5}
  \end{subfigure}
  \hfill
  \caption{B-MS vs. UCB base learners with confidence radius $c$ ($T= 10^3, R=100, K=5, M=6$).}
  \label{fig:ucb_results}
\end{figure}

\begin{figure}[H]
  \centering
  \begin{subfigure}[t]{0.45\linewidth}
    \centering
    \includegraphics[width=\linewidth]{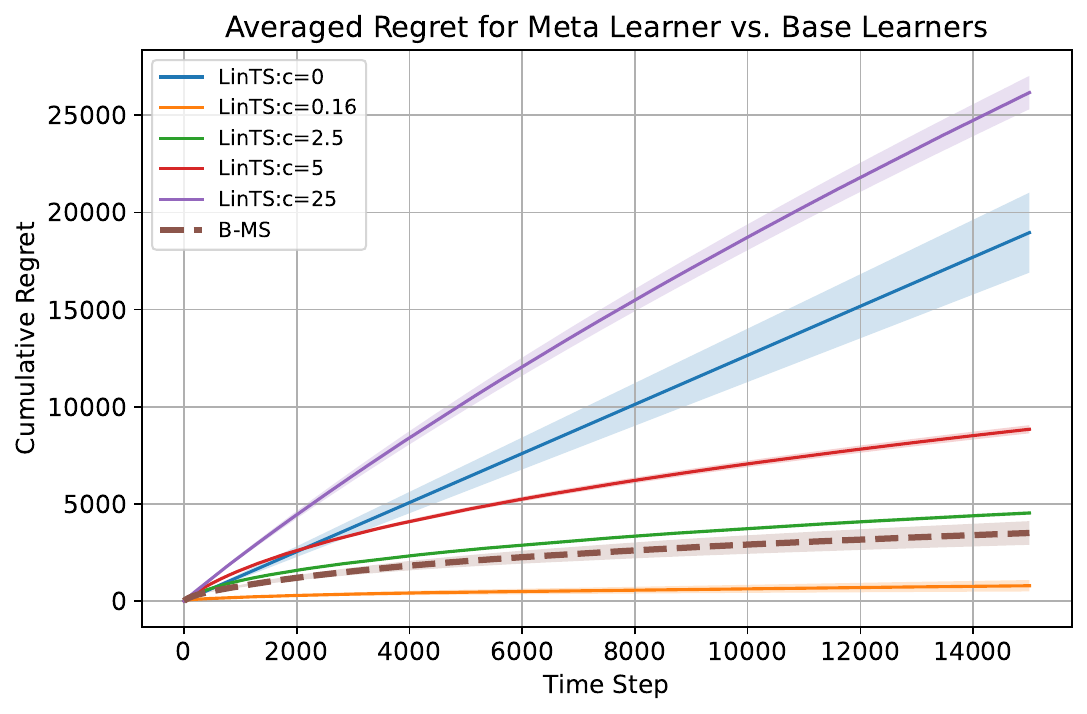}
    \caption{Empirical Bayesian regret}
    \label{fig:exp7-2_averaged_cum_regret_H15000_R100_v5}
  \end{subfigure}
  \hfill
  \begin{subfigure}[t]{0.5\linewidth}
    \centering
    \includegraphics[width=\linewidth]{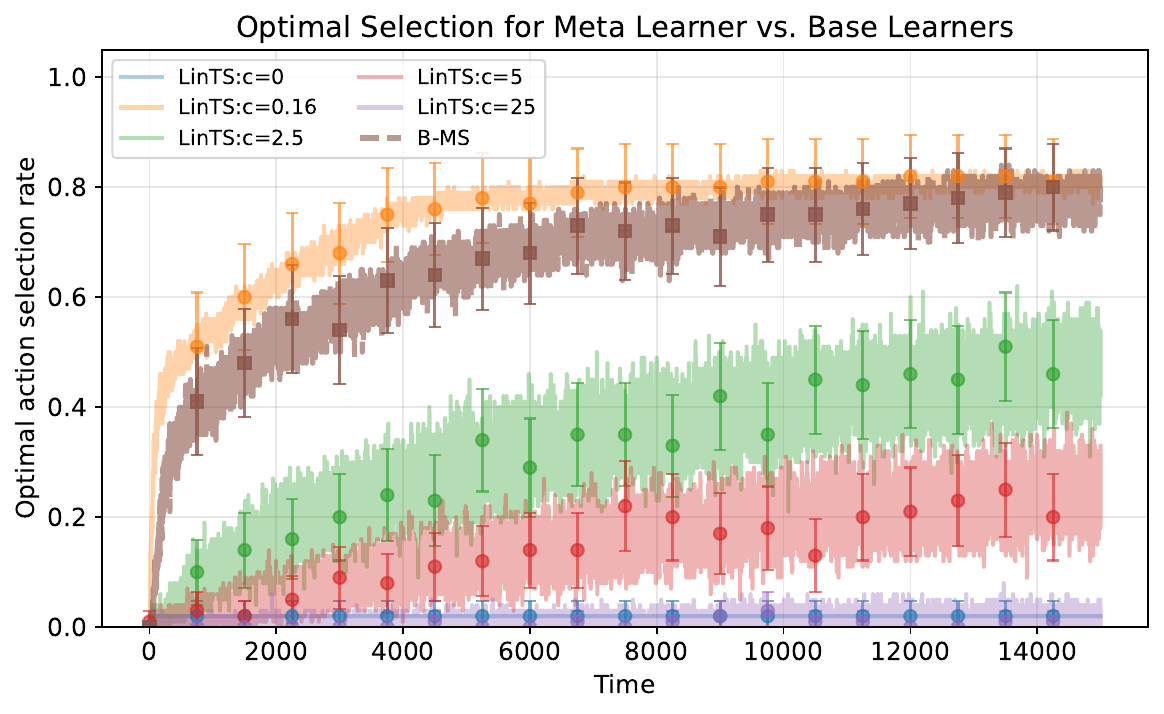}
    \caption{Optimal action selection rate}
    \label{fig:exp7-2_meta_statistics_opt_rate_H15000_R100_v5}
  \end{subfigure}
  \hfill
  \caption{B-MS vs. LinTS base learners with confidence radius $c$ ($T= 1.5\times 10^4, R=100, K=1000, M=5, d=10$).}
  \label{fig:lints_results}
\end{figure}

Additionally, our algorithm successfully avoids both under-exploration and over-exploration. In Figure \ref{fig:ucb_results}, the B-MS averaged regret curve demonstrates sublinear growth, contrasting with the approximately linear regret accumulation observed in nearly greedy strategies ($c=0.01$, $c=0.1$). Concurrently, it maintains substantially lower regret compared to configurations that over-explore ($c=5$, $c=10$). Figure \ref{fig:lints_results} further demonstrates that extreme parameter choices yield suboptimal behavior:
$c=0$ represents a purely exploitative strategy that rapidly accumulates high regret, while large values such as
$c=25$ induce excessive exploration, also resulting in poor performance. The meta learner successfully adapts across these configurations, matching the regret performance of the best LinTS variant.

We also investigated the effect of sharing data among base learners on the performance of the meta learner. The results are depicted in Figure \ref{fig:exp_all_share_vs_no}. We can see a common trend across all the plots that sharing data improves the performance of the meta learner and makes the algorithm more efficient. This is consistent with what we expect in theory, as sharing data allows each base learner to observe more samples, and therefore, advance faster through its learning curve. 

We further challenge our assumption on the well-specified prior \ref{subsec::assumption} in Figure \ref{fig:exp_all_share_vs_no}. We consider the case where the meta learner is mis-specified, but one of the base learners is well-specified (Figure \ref{fig:exp15}). Interestingly, we observe that even though the meta learner is mis-specified, the presence of one well-specified base learner in the pool helps the meta learner to recover from mis-specification. In the absence of any well-specified base learner, which is the setup of Figure \ref{fig:exp17}, a mis-specified meta learner would not show a competitive performance, reflecting the importance of the assumption on a well-specified prior.

\begin{figure}[H]
  \centering
  \begin{subfigure}[t]{0.45\linewidth}
    \centering
    \includegraphics[width=\linewidth]{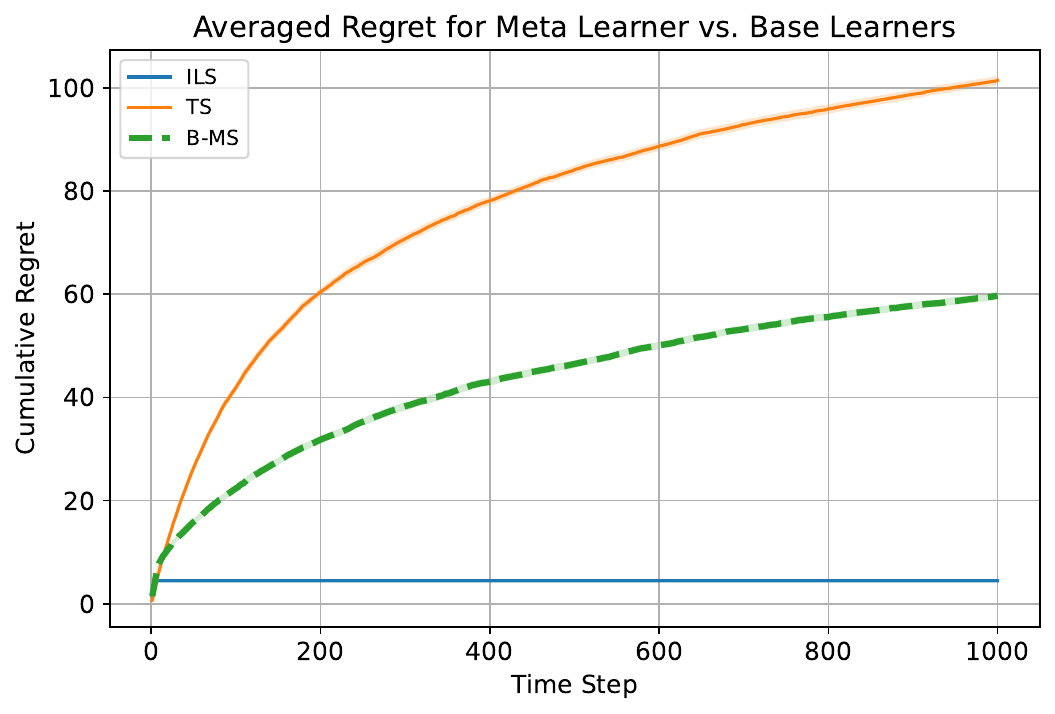}
    \caption{Empicical Bayesian regret}
    \label{fig:exp11_averaged_cum_regret_H1000_R60_v5}
  \end{subfigure}
  \hfill
  \begin{subfigure}[t]{0.5\linewidth}
    \centering
    \includegraphics[width=\linewidth]{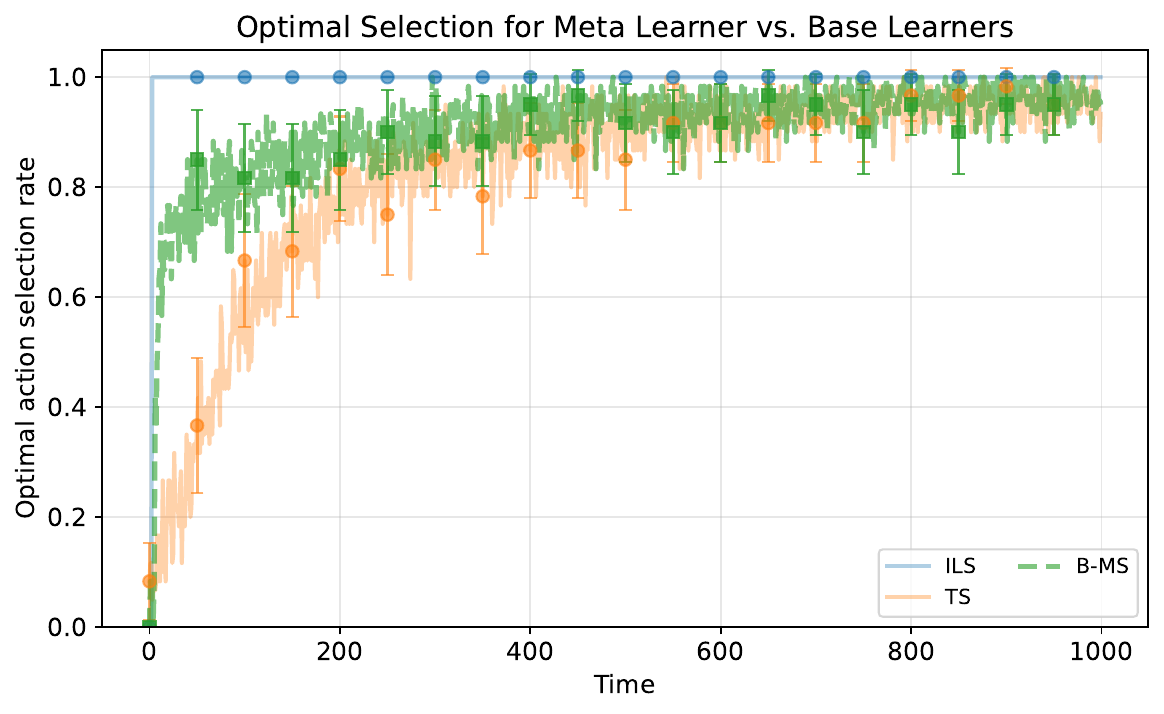}
    \caption{Optimal action selection rate}
    \label{fig:exp11_meta_statistics_opt_rate_H1000_R60_v5}
  \end{subfigure}
  \hfill
  \caption{B-MS vs. base learners (TS and ILS algorithm) for $T=10^3$, $R=60$, $K=8$, $M=3$.}
  \label{fig:ils_results}
\end{figure}

In the last experiment, we specify the base learners as a Thompson Sampling algorithm and a Information Lock Solver algorithm, and apply our proposed meta algorithm \ref{alg:modselTS}. As shown in Figure \ref{fig:exp11_averaged_cum_regret_H1000_R60_v5}, the meta learner successfully leverages diverse modeling assumptions of these base learners, achieving lower cumulative regret than the naive Thompson Sampling baseline. 
Importantly, the framework is not restricted to base learners that share the same algorithm but differ only by hyperparameter configurations. 

\section{Discussion and Future Work}
Our proposed algorithm works with a flexible online model selection framework in which any stochastic bandit algorithm can serve as a base learner. 
From a theoretical perspective, an interesting extension is to analyze the Bayesian regret lower bound of online model selection. As discussed in Section \ref{subsec::guarantee_discription}, prior work of \citep{marinov2021pareto} has established a lower bound of $\Omega(d{^{\star}}^2 \sqrt{T})$ for online model selection in the frequentist setting, but the Bayesian lower bound remains an open problem.

From an application perspective, the framework is particularly well-suited for online hyperparameter tuning, where multiple algorithms or parameter configurations must be evaluated adaptively under limited computational budgets. More broadly, our methodology captures applications beyond stochastic bandits, such as online learning scenarios with nonlinear reward structures, or reinforcement learning with structured priors, where different base learners encode distinct model assumptions. 

Following the development of the Diffusion Thompson Sampling and \citep{hsieh2023thompson, aouali2024diffusion}, an interesting direction is to leverage diffusion priors in Bayesian Online Model Selection. This extension would enable more expressive prior distributions that can capture a broader variation of model selection problems.

\section{Conclusion}
In this paper, we study the problem of online model selection in Bayesian stochastic bandits and introduce Bayesian Online Model Selection algorithm. Our selection strategy utilizes samples drawn from the posterior distribution to compute a potential function for each base learner. This potential function provides a randomized estimate of the realized regret that improves by posterior sampling.
 
Through careful analysis, we prove that the Bayesian regret of the algorithm enjoys an oracle-style guarantee of $\BRegret_T = \widetilde{\mathcal{O}} \left( d^\star M \sqrt{T} + \sqrt{MT} \right)$. We show that this method recovers the classic Thompson Sampling algorithm in its special case, and furthermore, has the advantage of leveraging structural information of the environment over Thompson Sampling.

We evaluate the empirical performance of our method on model selection tasks for multi-armed and linear bandit algorithms. Our algorithm is capable of tuning hyper-parameters of UCB and LinTS, achieving empirical performance that aligns with our theoretical analysis. 
Moreover, the experiments reflect that sharing data across base learners improves the overall performance of the algorithm, which is particularly helpful when the meta learner has mis-specified prior.


\bibliography{ref.bib}
\bibliographystyle{plainnat}

\newpage
\appendix

\section{Appendix}



\subsection{PROOFS}
\label{sec::appendix_proof}
\textit{In this section, we present the detailed proof of Lemmas and Theorems that we stated in the paper.}

\begin{corollary}
\label{corollary::well_specified_regret}
Denote $i_t^\star = \arg\min_{i \in [M]} \Regret_t^i$ as the base learner with minimum regret at time $t \in [T]$. Then,
\begin{align*}
         \Regret_t^{i_t^\star} \leq d^\star \sqrt{t}  \quad\quad \forall t \in [T]
\end{align*}
holds with probability at least $1 - \delta$.

\proof

Denote $j = \arg \min_{i \in [M]} d_T^i(\nu, \delta)$. By definition of the regret coefficient,
\begin{align*}
    \Regret_t^j \leq d_T^j(\nu, \delta) \sqrt{t}, \quad\quad \forall t \in [T]
\end{align*}

with probability at least $1 - \delta$. By definition of $d^{\star}$,
\begin{align*}
    d_T^j(\nu, \delta) \leq d^{\star}
\end{align*}

Therefore,
\begin{align*}
    \Regret_t^j \leq d_T^j(\nu, \delta) \sqrt{t} \leq d^{\star} \sqrt{t}, \quad\quad \forall t \in [T]
\end{align*}

Finally, by definition of $i_t^\star = \arg\min_{i \in [M]} \Regret_t^i$,
\begin{align*}
    \Regret_t^{i_t^{\star}} \leq \Regret_t^j \leq d^{\star} \sqrt{t},  \quad\quad \forall t \in [T]
\end{align*}
with probability at least $1 - \delta$.

\end{corollary}

\begin{lemma}
\label{lemma::dist_potentials}
Consider the probability space $(\Omega, \mathcal{F}, \mathbb{P})$ over the Bayesian world, and let $\mathcal{F}_{t}$ be the $\sigma$-algebra induced by all variables up to round $t$. Suppose the meta learner is interacting with env $\nu \sim \mathcal{P}$. Let $i_t^\star = \arg \min_{i \in [M]} \Regret_t^i$. The Bayesian Model Selection algorithm \ref{alg:modselTS} satisfies, 
\begin{align*}
    \mathbb{P} [ \arg\min_{i \in [M]} \Regret_t^i \mid \mathcal{F}_{t}] = \mathbb{P} [ \arg\min_{i \in [M]} \mathcal{\phi}_t^i \mid \mathcal{F}_{t}]  
\end{align*}
which implies,
\begin{align*}
      \mathbb{P} \left[ i_t = i \mid \mathcal{F}_{t} \right] = \mathbb{P} \left[ i_t^\star = i \mid \mathcal{F}_{t} \right]
\end{align*}

\proof

Recall the definition of the potential function,

\begin{align*}
    & \phi_t(i)[\tilde{\mu}_t] = n_t^i \tilde{\mu}_t^\star - \sum_{l \in \mathcal{I}_t^i} \tilde{\mu}_t(a_l) \\
    & \phi_t(i)[\mu] = n_t^i \mu^\star - \sum_{l \in \mathcal{I}_t^i} \mu_t(a_l) = \Regret_t^i.
\end{align*}

Conditioned on the observed history up to time $t$,

\begin{align*}
     \mathbb{P}[i_t = i \mid \mathcal{F}_{t}]  & = \mathbb{P}[\arg\min_{i \in [M]} \phi_t (i)[\tilde{\mu}_t] = i \mid \mathcal{F}_{t}] \tag{definition of $i_t$} \\
    & = \mathbb{P}[ \arg\min_{i \in [M]} \phi_t(i)[\mu] = i \mid \mathcal{F}_{t}]  \tag{correctness of posterior}\\
    & = \mathbb{P}[\arg\min_{i \in [M]} \Regret_t^i = i \mid \mathcal{F}_{t}]  \tag{definition of potential} \\
    & = \mathbb{P}[i^\star_t = i \mid \mathcal{F}_{t}] \tag{definition of $i_t^\star$}
\end{align*}

\end{lemma}

\begin{lemma}
\label{lemma::key_lemma_proof}
Denote $i_t^\star = \arg\min_{i \in [M]} \Regret_t^i$, and let $\mathcal{F}_{t}$ be the $\sigma$-algebra induced by all variables up to round $t$. For any function $f_t(i): \mathcal{[M]} \rightarrow \mathbb{R} $ that is $\mathcal{F}_{t}$-measurable,
\begin{align*}
    \mathbb{E} \left[ f_t(i_t^\star) \mid \mathcal{F}_{t} \right] = \mathbb{E} \left[ f_t(i_t) \mid \mathcal{F}_{t} \right]
\end{align*}

\proof 

\begin{align*}
    \mathbb{E} \left[ f_t(i_t) \mid \mathcal{F}_{t} \right] = \sum_{i=1}^M f_t(i) \, \mathbb{P} \left[ i_t = i \mid \mathcal{F}_{t} \right] 
\end{align*}

By lemma \ref{lemma::dist_potentials}, we have $\mathbb{P} \left[ i_t = i \mid \mathcal{F}_{t} \right] =  \mathbb{P} \left[ i_t^\star = i \mid \mathcal{F}_{t} \right]$, 
\begin{align*}
     = \sum_{i=1}^M f_t(i) \, \mathbb{P} \left[ i_t^\star = i \mid \mathcal{F}_{t} \right] = \mathbb{E} \left[ f_t(i_t^\star) \mid \mathcal{F}_{t} \right]
\end{align*}

\end{lemma}

\begin{lemma}[Good Event]
    \label{lemma::good_event_proof}
    Define the good event as, 
    \begin{align*}
        \mathcal{E}_{\text{good}} = \left\{ \left| u_t^i - \overline{u}_t^i \right| \leq c \sqrt{n_t^i \log\left(\frac{tM}{\delta}\right)} \quad\quad \forall i \in [M] \right\}
    \end{align*}
    
    The event $\mathcal{E}_{good}$ holds with probability at least $1 - \delta$.
\proof
By assumption 1 \ref{subsec::assumption}, the rewards are bounded in $[0, 1]$ and therefore the reward distributions are $\sigma$-subgaussian with variance proxy $\sigma^2 = \frac{1}{4}$. The statement follows by Hoeffding inequality and applying a union bound over $i \in [M]$.
\end{lemma}

\begin{lemma}
    \label{lemma::UCB_proof}
    Suppose $\mathcal{E}_{good}$ holds, and denote, 
    \begin{align*}
        f_t(i) = \frac{u_t^i}{n_t^i} +  c \sqrt{\frac{\log(tM/\delta)}{n_t^i}} + \frac{d^\star}{\sqrt{n_t^i}}
    \end{align*}
    For a base learner $i_t^\star = \arg\min_{i \in [M]} \Regret_t^i$, the function $f_t(i_t^\star)$ always overestimate the true mean $\mu^\star$,
    \begin{align*}
        \mu^\star \leq f_t(i_t^\star)
    \end{align*}

    \proof 
    By assumption \ref{subsec::assumption},
    \begin{align*}
        \Regret_t^{i_t^\star} = n_t^{i_t^\star} \mu^\star - \bar{u}_t^{i_t^\star} \, {\leq} \, d^\star \sqrt{n_t^{i_t^\star}}
    \end{align*}
    Rearrange,
    \begin{align*}
        n_t^{i_t^\star} \mu^\star & \leq \bar{u}_t^{i_t^\star} + d^\star \sqrt{n_t^{i_t^\star}} \\
     & \leq u_t^{i_t^\star} + c \sqrt{n_t^{i_t^\star} \log \left(\frac{tM}{\delta} \right)} + d^\star \sqrt{n_t^{i_t^\star}}  \tag{$\mathcal{E}_{\text{good}}$}    
    \end{align*}
    Divide both sides by $n_t^{i_t^\star}$,
    \begin{align*}
        \mu^\star \leq \frac{u_t^{i_t^\star}}{n_t^{i_t^\star}} + c \sqrt{\frac{\log \left(tM/\delta \right)}{ n_t^{i_t^\star}}} +  \frac{d^\star}{\sqrt{n_t^{i_t^\star}}} = f_t(i_t^\star)
    \end{align*}
\end{lemma}

\begin{lemma}\label{lemma:: average term upper bound_hp}
    For a fixed base learner $i \in [M]$,  denote,
    \begin{align*}
        g_t(i) = \left( \sum_{l \in \mathcal{I}_t^i} \frac{u_l^i}{n_l^i} \right) - u_t^i
    \end{align*}
For any $\delta \in (0,1)$ we have the following concentration bound,

\begin{equation*}
    \mathbb{P}_\nu\left(  g_{t}(i) \leq \phi_t(i)  + c\sqrt{n_t^i\log(TM/\delta)},\quad  ~\forall t \in [T] \right) \geq 1-T\delta
\end{equation*}

for all $t \in [T]$. Moreover, a union bound implies,
\begin{equation*}
    \mathbb{P}_\nu\left(  g_t(i) \leq  \phi_t(i)  + c\sqrt{n_t^{i}\log(TM/\delta)},~\forall i \in [M]   \text{ and } \forall t \in [T] \right) \geq 1-TM\delta
\end{equation*}
and therefore, 
\begin{equation*}
    \mathbb{P}_\nu\left(  g_t(i_t) \leq  \left( \min_{i} \phi_t(i)  \right) + c\sqrt{n_t^{i_t}\log(TM/\delta)} ~\forall t \in [T]  \right) \geq 1-TM\delta
\end{equation*}
where recall that $i_t = \min_i \phi_t(i)$. Finally,  
\begin{equation*}
\mathbb{P}_\nu\left(  g_{t(i)}(i) \leq  \phi_t(i_{t(i)})  + c\sqrt{n_t^{i_{t(i)}}\log(TM/\delta)}\right) \geq 1-TM\delta
\end{equation*}
where $t(i)$ is the last index in $[1,T]$ such that $i_{t(i)} = i$.

\proof

Let $\delta \in (0, 1)$. For all $\nu \in \text{Supp}(\mathcal{P})$, the following holds,
\begin{align*}
        g_t(i) &= \left(\sum_{l \in I_{t}^i} \frac{u_l^i}{n_l^i} \right) - u_{t}^i \\
        & \leq \left(\sum_{l \in I_{t}^i} \frac{\bar{u}_l^i}{n_l^i} + c\sqrt{\frac{\log (tM/ \delta )}{n_l^i}}\right) - u_{t}^i \tag{$\mathcal{E}_{\text{good}}$} \\
        & \leq \left(\sum_{l \in I_{t}^i} \frac{\bar{u}_l^i}{n_l^i} \right) - u_{t}^i + 2c \sqrt{n_{t}^i \log(tM /\delta)}  \\
        & \leq \left(\sum_{l \in I_{t}^i} \frac{\bar{u}_l^i}{n_l^i} \right) - \bar{u}_{t}^i + 3c \sqrt{n_{t}^i \log(tM /\delta)} \tag{$\mathcal{E}_{\text{good}}$} \\
        & \leq n_{t}^i \mu^\star - \bar{u}_t^i \tag{$\frac{\bar{u}_l^i}{n_l^i} \leq \mu^\star$} +  3c \sqrt{n_{t}^i \log(tM /\delta)} \\
        & \leq \Regret_{t-1}^i + 3c \sqrt{n_{t}^i \log(tM /\delta)} \tag{Definition of Regret} \\
        & \leq \Regret_{t}^i + 3c \sqrt{n_{t}^i \log(tM /\delta)} \tag{Monotonicity of Regret}
    \end{align*}
    
with probability at least $1-\delta$, for any $t \in [T]$ and $i \in [M]$. This statement can be written as follows, 
\begin{equation*}
    \mathbb{P}\left( g_t(i)  \leq \Regret_{t}^i + 3c \sqrt{n_{t}^i \log(tM /\delta)} ~\Big |~ \nu \right) \geq 1-\delta
\end{equation*}
for all environments $\nu$ and therefore, by marginalizing it also follows, 
\begin{equation*}
   \mathbf{I} =  \mathbb{P}_\nu\left(  g_t(i)  \leq \Regret_{t}^i + 3c \sqrt{n_t^i \log(tM /\delta)} \right) \geq 1-\delta. 
\end{equation*}
Term $\mathbf{I}$ can be written as the following integral,
\begin{equation*}
 \mathbf{I} =    \int_{\mathcal{H}_t} \mathbb{P}_\nu\left(  g_t(i)  \leq \Regret_t^i + 3c \sqrt{n_t^i \log(tM /\delta)}    \Big | \mathcal{H}_t\right)\mathbb{P}( \mathcal{H}_t) \mathrm{d}_{\mathcal{H}_t}
\end{equation*}
where $\mathcal{H}_t$ corresponds to histories up to time $t$. By definition of the potential, the following equation holds, 
$$\mathbb{P}_\nu\left( g_t(i)  \leq \Regret_t^i + 3c \sqrt{n_t^i \log(tM /\delta)}    \Big | \mathcal{H}_t\right) = \mathbb{P}_\nu\left(  g_t(i)  \leq \phi_t (i) + 3c \sqrt{n_t^i \log(tM /\delta)}    \Big | \mathcal{H}_t\right)$$

Substituting back, we get,

\begin{equation*}
 \mathbf{I} = \int_{\mathcal{H}_t} \mathbb{P}_\nu\left(  g_t(i) \leq \phi_t(i) + 3c \sqrt{n_t^i \log(tM /\delta)}   \Big | \mathcal{H}_t\right)\mathbb{P}( \mathcal{H}_t) \mathrm{d}_{\mathcal{H}_t} \geq 1-\delta
\end{equation*}

Therefore,
\begin{equation*}
  \mathbb{P}_\nu\left( g_t(i)  \leq \phi_t(i) + 3c \sqrt{n_t^i \log(tM /\delta)} \right) \geq 1-\delta
\end{equation*}

Applying union bound over $t \in [T]$, 
\begin{equation*}
    \mathbb{P}_\nu\left(  g_t(i) \leq \phi_t(i)  + c\sqrt{n_t^i\log(MT/\delta)}   ~\forall t \in [T] \right) \geq 1-T\delta
\end{equation*}

Applying union bound over $i \in [M]$,
\begin{equation*}
    \mathbb{P}_\nu\left( g_t(i) \leq  \phi_t(i)  + c\sqrt{n_t^{i}\log(MT/\delta)}, ~\forall i \in [M]   \text{ and } \forall t \in [T] \right) \geq 1-MT\delta
\end{equation*}

Finally, this implies,

\begin{equation*}
\mathbb{P}_\nu\left(  g_t(i_t) \leq  \phi_t(i_t)  + c\sqrt{n_t^{i_t}\log(MT/\delta)}, ~\forall t \in [T] \right) \geq 1-MT\delta
\end{equation*}

and in particular,
\begin{equation*}
\mathbb{P}_\nu\left(  g_{t(i)}(i) \leq  \phi_{t(i)}(i)  + c\sqrt{n_{t(i)}^{i}\log(MT/\delta)}\right) \geq 1-MT\delta
\end{equation*}

where $t(i)$ is the last index in $[T]$ such that $i_{t(i)} = i$. 

\end{lemma}

\begin{lemma}
\label{lemma::potential_upperbound_hp_proof}
The following inequality holds,
\begin{equation*}
    \mathbb{P}_\nu\left( \phi_t(i_t) \leq d^\star\sqrt{t}, \quad ~\forall t \in [T] \right) \geq 1-T\delta
\end{equation*}

\proof

By corollary \ref{corollary::well_specified_regret} for any $\nu \sim \mathcal{P}$, with probability at least $1-\delta$,
\begin{equation*}
    \Regret_{t}^{i_t^\star}  \leq d^\star \sqrt{t}, \quad\quad \forall t \in [T]
\end{equation*}

Thus, we have, 
\begin{equation*}
 \mathbb{P}\left(    \Regret_{t}^{i_t^\star} \leq d^\star \sqrt{t},  \quad ~\forall t \in [T]~\Big| \nu \right) \geq 1-\delta.
\end{equation*}

Therefore, by marginalizing,
\begin{equation*}
\mathbf{I} = \mathbb{P}_\nu\left(    \Regret_{t}^{i_t^\star} \leq d^\star \sqrt{t},  \quad ~\forall t \in [T] \right) \geq 1-\delta.
\end{equation*}

Term $\mathbf{I}$ evaluated at a single $t$ can be written as the following integral,
\begin{equation*}
    \int_{\mathcal{H}_t} \mathbb{P}_\nu\left(  \Regret_{t}^{i_t^\star} \leq d^\star \sqrt{t }   \Big | \mathcal{H}_t\right)\mathbb{P}( \mathcal{H}_t) \mathrm{d}_{\mathcal{H}_t}  \geq 1-\delta
\end{equation*}

By the correctness of the posterior,
\begin{equation*}
    \int_{\mathcal{H}_t} \mathbb{P}_\nu\left(  \phi_{t}(i_t^\star) \leq d^\star \sqrt{t }   \Big | \mathcal{H}_t\right)\mathbb{P}( \mathcal{H}_t) \mathrm{d}_{\mathcal{H}_t}  \geq 1-\delta
\end{equation*}
Now since $\phi_t(i)$ and $d^\star \sqrt{t} $ are $\mathcal{F}_{t}$-measurable, we can apply lemma \ref{lemma::key_lemma}, 
\begin{equation*}
    \int_{\mathcal{H}_t} \mathbb{P}_\nu\left(  \phi_{t}(i_t) \leq d^\star \sqrt{t }   \Big | \mathcal{H}_t\right)\mathbb{P}( \mathcal{H}_t) \mathrm{d}_{\mathcal{H}_t}  \geq 1-\delta
\end{equation*}

Therefore,
\begin{equation*}
    \mathbb{P}_\nu\left(  \phi_{t}(i_t) \leq d^\star \sqrt{t }   \right)  \geq 1-\delta
\end{equation*}
the result then follows via a union bound over $t \in[T]$.
\end{lemma}

\begin{corollary}
\label{corollary::sum of average terms}
For any $t \in [T]$, denote, $$g_t(i) = \sum_{l \in I_t^i} \frac{u_l^i}{n_l^i} - u_t^i$$Then, the following holds,
\begin{equation*}
\mathbb{E}_\nu \left[ \sum_{i \in [M]} g_T(i) \right] \leq d^\star M \sqrt{T} + c \sqrt{TM \log\left(\frac{TM}{\delta} \right)} + 2T^2M^2 \delta
\end{equation*}

\proof
Denote $t(i) \in [T]$ as the last timestep that base learner $i \in [M]$ was selected. Since $i$ was not selected between times $t(i)$ and $T$, we have $g_T(i) = g_{t(i)}(i)$. Therefore,

\begin{align*}
    \mathbb{E}_\nu \left[ \sum_{i \in [M]} g_T(i) \right] & = \mathbb{E}_\nu \left[ \sum_{i \in [M]} g_{t(i)}(i) \right] \\
    & \leq \mathbb{E}_\nu \left[ \sum_{i \in [M]} \phi_{t(i)}(i) + c \sqrt{n_t^i \log\left(\frac{TM}{\delta} \right) }\right] \tag{Lemma \ref{lemma:: average term upper bound_hp}} + T^2M^2 \delta \\
    & \leq  \mathbb{E}_\nu \left[ \sum_{i \in [M]} \phi_{t(i)}(i)  +  c\sqrt{n_T^i \log\left(\frac{TM}{\delta} \right)} \right] + T^2M^2 \delta \tag{$n_t^i \leq n_T^i$} \\
    & \leq  \mathbb{E}_\nu \left[ \sum_{i \in [M]} \phi_{t(i)}(i) \right]  +  c\sqrt{TM \log\left(\frac{TM}{\delta} \right)}  + T^2M^2 \delta \tag{Cauchy-Schwarz} \\
    & \leq  \mathbb{E}_\nu \left[ \sum_{i \in [M]} \max_{t \leq T} \phi_{t}(i_t) \right] +  c\sqrt{TM \log\left(\frac{TM}{\delta} \right)}  + T^2M^2 \delta \tag{max} \\
    & = M \, \mathbb{E}_{\nu} \left[ \max_{t \leq T} \phi_{t}(i_t) \right] + c\sqrt{TM \log\left(\frac{TM}{\delta} \right)}  + T^2M^2 \delta \\
    & \leq d^\star M \sqrt{T} + c \sqrt{TM \log\left(\frac{TM}{\delta} \right)} + 2T^2M^2 \delta
    \tag{Lemma \ref{lemma::potential_upperbound_hp_proof}}
\end{align*}

\end{corollary}

\begin{theorem}[Oracle-Best Guarantee]
\label{theorem::Bayesian_Regret_Bound_proof}

The Bayesian Regret of Algorithm \ref{alg:modselTS} satisfies, 

\begin{align*}
    \BRegret_T \leq  \tilde{\mathcal{O}}\left(  d^\star M \sqrt{T} + \sqrt{MT} \right) 
\end{align*}
where the big $\tilde{\mathcal{O}}$ hides the constants and log factors.

\proof
Denote, 
\begin{align*}
    f_t(i) = \frac{u_t^i}{n_t^i} +  c \sqrt{\frac{\log(tM/\delta)}{n_t^i}} + \frac{d^\star}{\sqrt{n_t^i}}
\end{align*}

Suppose $\mathcal{E}_{good}$ holds,
\begin{align*}
    \BRegret_T  = \mathbb{E}_{\nu} \left[ \sum_{t=1}^T \mu^\star  - \mu(a_t)\right] = \mathbb{E}_{\nu} \left[ \sum_{i=1}^M \sum_{t \in I_T^i} \mu^\star - \mu^i_t\right]  
\end{align*}

Add and Subtract $f_t(i)$,
\begin{align*}
    = \mathbb{E}_{\nu} \left[ \sum_{i=1}^M \sum_{t \in I_T^i} \mu^\star - f_t(i) + f_t(i) - \mu^i_t\right]
\end{align*}
\begin{align*}
    = \mathbb{E}_{\nu} \left[  \sum_{i=1}^M \sum_{t \in I_T^i}  \mu^\star - f_t(i) \right] + \mathbb{E}_{\nu} \left[  \sum_{i=1}^M \sum_{t \in I_T^i} f_t(i) - \mu_t^i \right] 
\end{align*}
\begin{align*}
    = \mathbb{E}_{\nu} \left[  \sum_{i=1}^M \sum_{t \in I_T^i} \mathbb{E} \left[ \mu^\star - f_t(i) \mid \mathcal{F}_{t} \right] \right] + \mathbb{E}_{\nu} \left[  \sum_{i=1}^M \sum_{t \in I_T^i} f_t(i) - \mu_t^i \right] 
\end{align*}

By Lemma \ref{lemma::key_lemma}, $\mathbb{E} \left[ f_t(i_t) \mid \mathcal{F}_{t} \right] = \mathbb{E} \left[ f_t(i_t^\star) \mid \mathcal{F}_{t} \right] $, therefore,
\begin{align*}
    = \mathbb{E}_{\nu} \left[  \sum_{i=1}^M \sum_{t \in I_T^i} \mathbb{E} \left[ \mu^\star - f_t(i_t^\star)  \mid \mathcal{F}_{t} \right] \right] + \mathbb{E}_{\nu} \left[  \sum_{i=1}^M \sum_{t \in I_T^i} f_t(i) - \mu_t^i \right] 
\end{align*}

By lemma \ref{lemma::UCB}, $\mu^\star \leq f_t(i_t^\star)$ for all $t \in [T]$. Therefore, the first term is negative,

\begin{align*}
    \leq \mathbb{E}_{\nu} \left[ \sum_{i=1}^M \sum_{t \in I_T^i} f_t(i) - \mu_t^i\right]
\end{align*}

Substitute definition of $f_t(i_t)$,

\begin{align*}
    = \mathbb{E}_{\nu} \left[  \sum_{i=1}^M \sum_{t \in I_T^i} \frac{u_t^i}{n_t^i} +  c \sqrt{\frac{\log(tM/\delta)}{n_t^i}} + \frac{d^\star}{\sqrt{n_t^i}} - \mu_t^i \right]
\end{align*}

Rearrange, 
\begin{align*}
    = \mathbb{E}_{\nu} \left[  \sum_{i=1}^M \sum_{t \in I_T^i} \frac{u_t^i}{n_t^i} - \mu_t^i  \right] +   \mathbb{E}_{\nu} \left[  \sum_{i=1}^M \sum_{t \in I_T^i} c \sqrt{\frac{\log(tM/\delta)}{n_t^i}} + \frac{d^\star}{\sqrt{n_t^i}}  \right]
\end{align*}

\begin{align*}
    \leq \mathbb{E}_{\nu} \left[ \sum_{i=1}^M \sum_{t \in I_T^i} \frac{u_t^i}{n_t^i} - \mu_t^i  \right] +   \mathbb{E}_{\nu} \left[ \left(d^\star + c \sqrt{\log(TM/\delta)}   \right) \sum_{i=1}^M \sum_{t \in I_T^i} \frac{1}{\sqrt{n_t^i}} \right]
\end{align*}

The term $\sum_{t=1}^{T} \frac{1}{\sqrt{n_t^i}}$ can be bounded as follows,

\begin{align*}
    \sum_{i=1}^{M} \sum_{l \in \mathcal{I}_t^i}^{n_t^i} \frac{1}{\sqrt{n_l^i}} \leq \sum_{i=1}^{M} \sum_{\ell = 1}^{n_t^i} \frac{1}{\sqrt{\ell}} \leq \mathcal{O} \left( \sum_{i=1}^{M} \sqrt{n_t^i}\right)
\end{align*}

By Cauchy-Schwarz,
\begin{align*}
    \leq \mathcal{O} \left( \sqrt{ M \sum_{i=1}^{M} n_t^i }\right) \leq \mathcal{O} \left( \sqrt{MT} \right)
\end{align*}

By substituting this bound, we get, 
\begin{align*}
    \mathbb{E}_{\nu} \left[ \sum_{i=1}^M \sum_{t \in I_T^i} \mu^\star  - \mu^i_t \right]
    \leq \mathbb{E}_{\nu} \left[ \sum_{i=1}^M \sum_{t \in I_T^i} \frac{u_t^i}{n_t^i} - \mu_t^i  \right] +  \mathcal{O} \left(d^\star\sqrt{MT}  + c \sqrt{MT \log \left(\frac{MT}{\delta} \right)} \right)
\end{align*}

It remains to bound the first term. 

\begin{align*}
    \mathbb{E}_{\nu} \left[ \sum_{i=1}^M \sum_{t \in I_T^i}  \frac{u_t^i}{n_t^i} - \mu_t^i  \right] & = \mathbb{E}_{\nu} \left[ \sum_{i=1}^M \left( \sum_{t \in I_T^i} \frac{u_t^i}{n_t^i} \right) - \bar{u}_T^i \right]  \\
    & \leq  \mathbb{E}_{\nu} \left[ \sum_{i=1}^M  \left( \sum_{t \in I_T^i} \frac{u_t^i}{n_t^i} \right) - u_T^i + c \sqrt{n_T^i \log \left(\frac{MT}{\delta} \right)} \right] \tag{$\mathcal{E}_{\text{good}}$} \\
    & \leq  \mathbb{E}_{\nu} \left[ \sum_{i=1}^M \left( \sum_{t \in I_T^i} \frac{u_t^i}{n_t^i} \right) - u_T^i \right] + \mathcal{O} \left( c \sqrt{MT \log \left(\frac{MT}{\delta} \right)} \right)
\end{align*}

Denote and substitute $g_t(i) = \sum_{l \in \mathcal{I}_t^i} \frac{u_l^i}{n_l^i} - u_t^i$,

\begin{align*}
    & = \mathbb{E}_\nu \left[ \sum_{i \in [M]} g_T(i) \right] +  2c \sqrt{MT \log \left(\frac{MT}{\delta} \right)}  \\
\end{align*}

 By corollary \ref{corollary::sum of average terms},

\begin{align*}
    & \leq d^\star M \sqrt{T}  +  3c \sqrt{MT \log \left(\frac{MT}{\delta} \right)} + 2M^2T^2 \delta
\end{align*}

Combine all terms, and set $\delta = \frac{1}{M^2T^2}$,

\begin{align*}
    \BRegret (T) &= \mathcal{O}\left( (d^\star \sqrt{MT})(1+ \sqrt{M}) + \sqrt{MT \log(MT)}\right) \\
    & = \mathcal{O}\left( d^\star M\sqrt{T} + \sqrt{MT}\right)
\end{align*}

\end{theorem}

\begin{theorem}
\label{theorem::recovering_TS_proof}

Consider a $k$-armed bandit environment. Suppose we use Algorithm \ref{alg:modselTS} with $k$ base learners, where base learner $i$ only pulls arm $i \in [k]$ for $t=[1, \cdots, T]$. Then,  Algorithm \ref{alg:modselTS} recovers the bound of Thompson Sampling by achieving, 

\begin{align*}
    \BRegret_T \leq  \tilde{\mathcal{O}}\left( \sqrt{KT} \right) 
\end{align*}

\proof 

Denote $[\mu_1, \cdots, \mu_k]$ as the vector of mean rewards, and $i^\star = \arg\max_{i \in [M]} \mu^i$ as the optimal arm. Since the $i$'th base learner always pulls arm $i \in[k]$, then base learner $i^\star \in [k]$ always pulls the optimal arm, and hence achieves zero regret. By definition \ref{def::BRC} of the regret coefficient,  $d^\star = 0$, and hence Algorithm \ref{alg:modselTS} achieves,
\begin{align*}
    \BRegret_T = \tilde{\mathcal{O}} \left( \sqrt{KT} \right)
\end{align*}

This bound is equal to the known upper bound on the Bayesian regret of Thompson Sampling up to log factors. \citep{Lattimore_Szepesvári_2020}.

\end{theorem}

\subsection{REMARKS ABOUT ASSUMPTIONS}
\label{appe::assumption_remarks}
Our assumptions in Section \ref{subsec::assumption} are minimal and standard in the literature of online model selection and Thompson Sampling to derive meaningful regret guarantees. Here are some remarks to address the cases where these assumptions might not hold.
\begin{enumerate}
    \item \textbf{(Boundedness)} We assume the rewards are bounded $r \in [0, 1]$ at all timesteps. For the more general case of $r \in [-R, R]$, one can always normalize the reward and repeat our analysis. The final bounds would be equal up to constant factors. 
    \item \textbf{(Well-Specified Prior)} Our analysis requires the meta learner to have access to a correct prior. In experiments of Figure \ref{fig:exp_all_share_vs_no}, we discussed that if this assumption is violated, the presence of one well-specified base learner would help the meta learner to recover from mis-specification. 
    
    Additionally, prior work has proposed solutions to handle misspecified priors. For stochastic bandits, \citep{kveton2021meta} designed a meta learning algorithm to learn correct priors through interaction. For the more general decision making setting, \citep{hong_thompson_2022} suggests the use of mixture priors.

\end{enumerate}

\subsection{BASE ALGORITHMS}\label{appe:algos}

Each base learner maintains its own sufficient statistics, such as the cumulative reward and the number of times each action has been played. These statistics are updated only when the corresponding base learner is selected by the meta learner. At global round \(t\), when the meta learner selects base learner \(i_t\), it executes the action recommended by \(i_t\) and observes the resulting reward. For example, if the selected base learner implements the UCB strategy, we execute Algorithm~\ref{alg: mab-ucb}; if it implements the linear bandit model, we run Algorithm~\ref{alg: linTS}. We define $T_b$ is batch horizon. For non-batched version, we assume $T_b=1$. In all experiments, we set the batch horizon to \(T_b = 1\), so that base learners are updated once at every selection.

\begin{algorithm}
    \caption{UCB for Multi-armed Bandit}
    \label{alg: mab-ucb}
    \begin{algorithmic}
        \REQUIRE Base learner index $i$, batch horizon $T_b$, initial reward sum $u_1^{i}$ and action selection count $n_1^{i}$.
        \FOR{$\tau=1, ..., T_{b}$}
            \STATE Compute empirical mean $\hat{\mu}_{\tau}^i(a)=\frac{u_{\tau}^i(a)}{n_{\tau}^i(a)}$
            \STATE Compute for all $a\in \mathcal{A}$ 
            $$\text{UCB}_\tau(a) = \hat{\mu}_{\tau}^i(a) + c_i\sqrt{\frac{\log{(2\cdot \sum_{a\in\mathcal{A}}n_\tau^i(a)K)/\delta)}}{n_\tau^i(a)}}$$ 
            \STATE Select action $a_\tau = \arg\max_{a\in\mathcal{A}}\text{UCB}_\tau(a)$
            \STATE Receive reward $r_\tau$
            \STATE Compute statistics using all action and reward data given base learner $i$
            $$n_{\tau+1}^i(a) = n_\tau^i(a) + 1 \quad \text{and} \quad 
            u_{\tau+1}^i(a) = u_\tau^i(a) + r_{\tau}$$
        \ENDFOR
        \STATE Return $n_{T_{b}}^i$, $u_{T_{b}}^i$
    \end{algorithmic}
\end{algorithm}

\begin{algorithm}
    \caption{Linear Thompson Sampling}
    \begin{algorithmic}
        \REQUIRE Base learner index $i$, batch horizon $T_b$, dimension of action space $d$, initial statistics $\hat{\theta}_{1}^i$, $V_{1}^i$
        \FOR{$\tau = 1, \dots, T_b$}
            \STATE Sample $\tilde{\theta}_{\tau}^{i} \sim \mathcal{N}(\hat{\theta}_{\tau}^i, c_i^2d V_{\tau}^{i,-1})$
            \STATE Select arm $a_\tau = \arg\max_{a\in\mathcal{A}} \left\langle \tilde{\theta}_{\tau}^{i},  a\right\rangle$
            \STATE Pull arm $a_\tau$ and observe reward $r_\tau$
            \STATE Compute statistics using all action and reward data given base learner $i$
            $$V_{\tau+1}^i = V_{\tau}^i + a_{\tau}a_\tau^\top \quad \text{and} \quad \hat{\theta}_{\tau+1}^i = V_{\tau+1}^{i,-1}(\hat{\theta}_{\tau-1}V_{\tau}^i+a_\tau r_{\tau})$$
        \ENDFOR
    \STATE Return $V_{T_b}^i$, $\hat{\theta}_{T_b}^i$
    \end{algorithmic}
    \label{alg: linTS}
\end{algorithm}

\subsection{CONNECTION TO THOMPSON SAMPLING WITH EXPERIMENT}
We consider the same bandit environment as in the previous experiment, but modify the setup so that each base learner selects one action continuously over time. In this case, the number of base learners coincides with the number of actions. The policy of base learner $i\in[M]$ is
\[
a_t = i
\]
Consequently, the regret and the regret coefficient of the optimal base learner, which selects the optimal arm all the time, is zero, which reduces the Bayesian regret of the meta algorithm to $\sqrt{MT} = \sqrt{KT}$, matching the regret rate of Thompson Sampling.

\begin{figure}[H]
    \centering
    \includegraphics[width=0.45\linewidth]{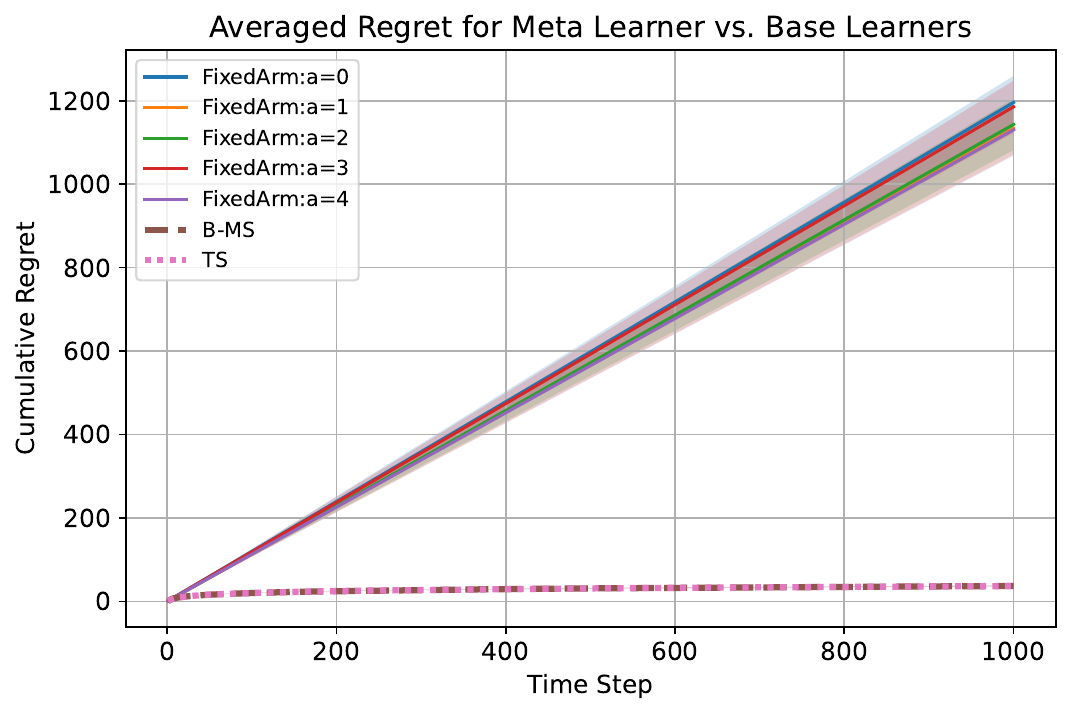}
    \caption{Empirical Bayesian regret for B-MS vs. TS and Fixed Arm base learners with action index $a$ ($T= 10^3, R=10^3, K=M=5$)}
    \label{fig:exp3_averaged_cum_regret_H1000_R1000_v5}
\end{figure}

As mentioned in Section \ref{sec:connection2ts}, shown in Figure~\ref{fig:exp3_averaged_cum_regret_H1000_R1000_v5}, the regret curve of our method aligns almost exactly with that of Thompson Sampling and attains the lowest regret under this scenario. To be noted, the remaining base learners exhibit approximately linear regret, since different bandit instances sampled from the Bayesian environment induce different action orderings and no single learner is optimal across all instances. As the number of runs increases, the averaged regret curves are therefore expected to overlap.

\subsection{WELL-SPECIFICATION \& MIS-SPECIFICATION}

In order to discuss the effect of prior specification, we want to use Thompson Sampling as base learner as an example for illustration as talked about in Section \ref{sec:experiment}. There are four cases,
\begin{itemize}
    \item Both the meta-prior and at least one base learner’s prior are correctly specified.
    \item A well-specified meta-prior exists, whereas all base learners are associated with mis-specified priors.
    \item The meta-prior is mis-specified, while at least one base learner has a well-specified prior.
    \item Both the meta-prior and the priors of all base learners are mis-specified.
\end{itemize}

\begin{figure}[htbp]
  \centering

  \begin{subfigure}[t]{0.45\linewidth}
    \centering
    \includegraphics[width=\linewidth]{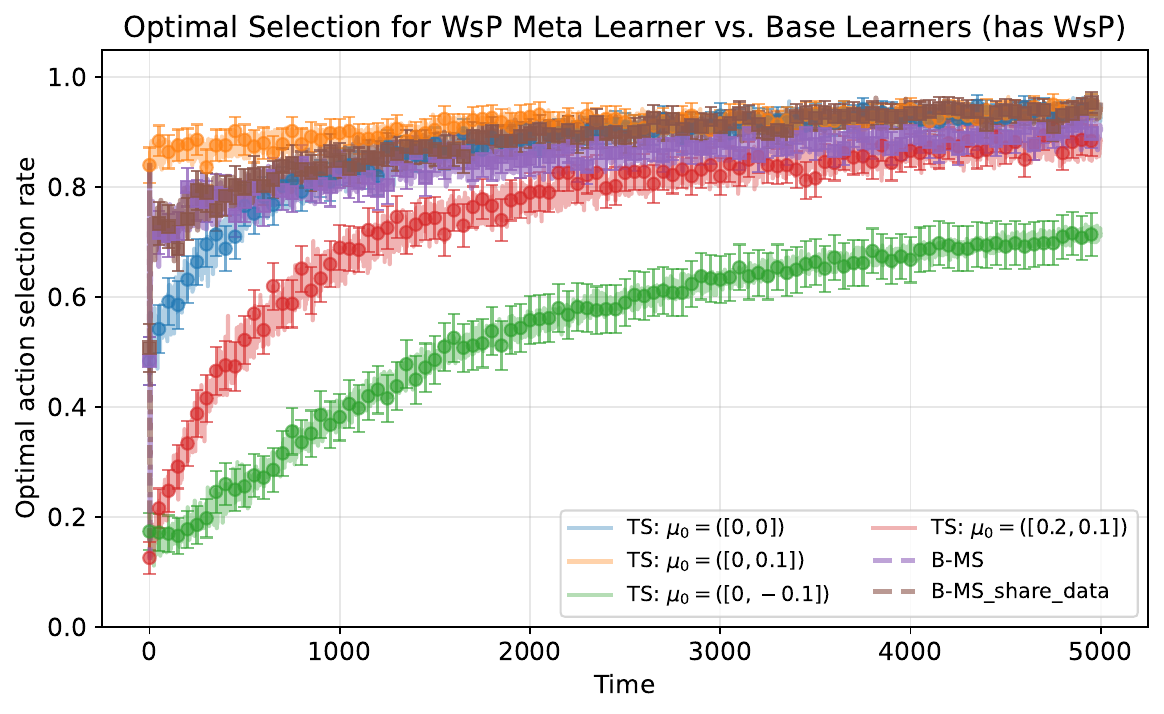}
    \caption{Exp.~14: Well-specified meta learner and one base learner}
    \label{fig:exp14_action}
  \end{subfigure}
  \hfill
  \begin{subfigure}[t]{0.45\linewidth}
    \centering
    \includegraphics[width=\linewidth]{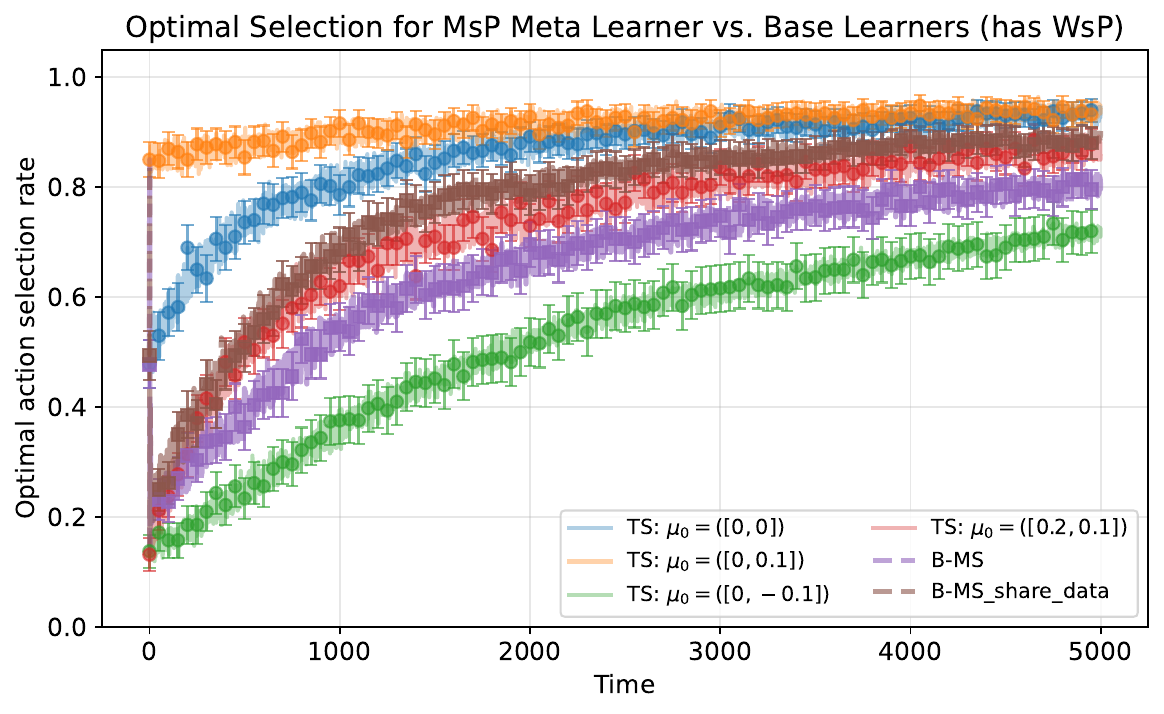}
    \caption{Exp.~15: Mis-specified meta learner and one well-specified base learner}
    \label{fig:exp15_action}
  \end{subfigure}
  \begin{subfigure}[t]{0.45\linewidth}
    \centering
    \includegraphics[width=\linewidth]{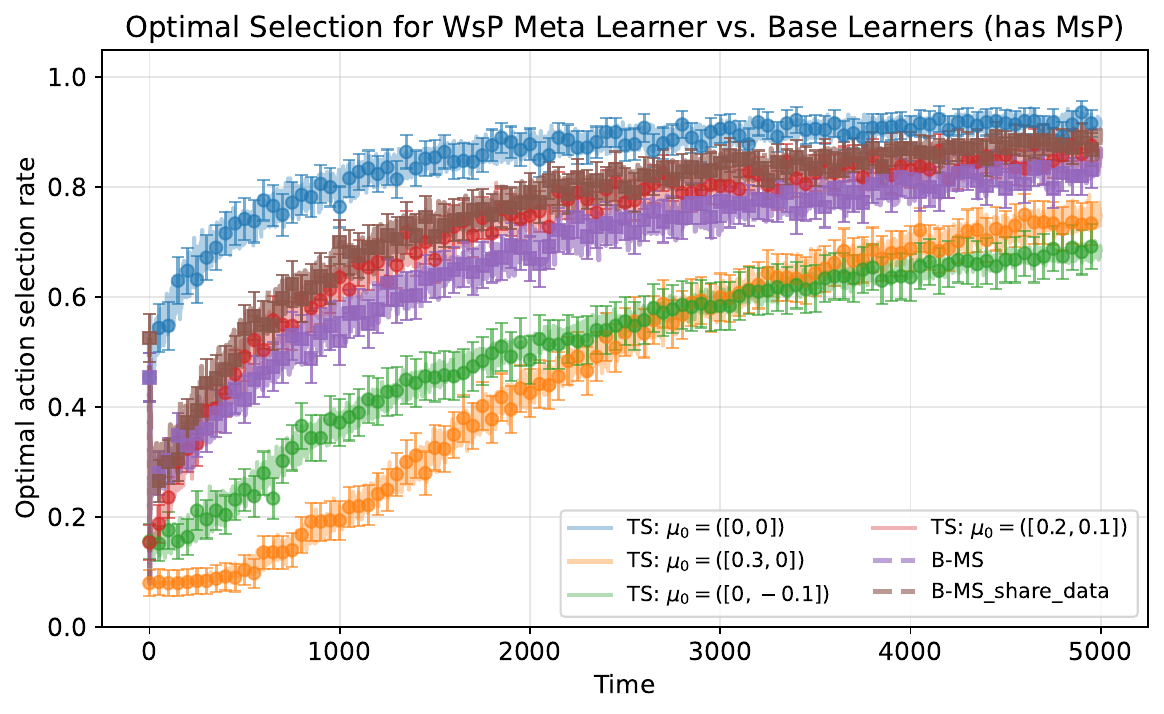}
    \caption{Exp.~16: Well-specified meta learner and all mis-specified base learners}
    \label{fig:exp16_action}
  \end{subfigure}
  \hfill
  \begin{subfigure}[t]{0.45\linewidth}
    \centering
    \includegraphics[width=\linewidth]{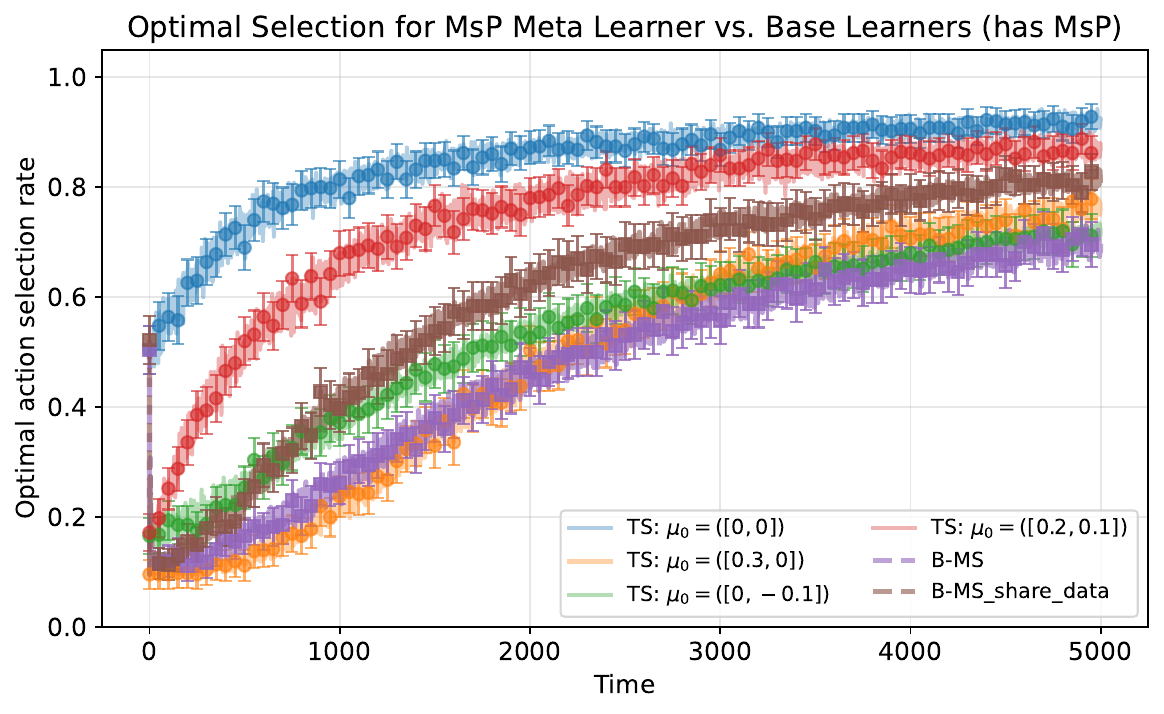}
    \caption{Exp.~17: Mis-specified meta learner and base learners}
    \label{fig:exp17_action}
  \end{subfigure}
  \hfill
  \caption{Optimal action selection rate for B-MS vs. TS base learners with prior specifications ($T=5\times10^3$, $R=500$, $K=2$, $M=4$)}
  \label{fig:exp_all_share_vs_no_action}
\end{figure}

The posterior distribution of each arm can be written as for arm $i\in \{1, 2\}$
$$
\mathcal{P}_t^i = \mathcal{N}\left(\frac{\sum_{l=1}^{n_t^i}r_l+\mu_0/\sigma_0^2}{n_t^i+1/\sigma_0^2}, \frac{1}{n_t^i+1/\sigma_0^2}\right)
$$
Consider posterior mean as weighted averaged between prior mean $\mu_0$ and sampled mean $\bar{y} = \frac{1}{n_t^i}\sum_{l=1}^{n_t^i}r_l$, 
\[
\underbrace{\mu_n}_{\text{posterior mean}}
= 
\frac{\frac{1}{\sigma_0^2}}{\frac{1}{\sigma_0^2} + n}
\underbrace{\mu_0}_{\text{prior mean}}
+
\frac{n}{\frac{1}{\sigma_0^2} + n}
\underbrace{\bar{y}}_{\text{sample mean}} \]
If
\(
n<\frac{1}{\sigma_0^2}
\), the posterior mean relies more on prior mean specification;
if $n>\frac{1}{\sigma_0^2}$, data will affect posterior more. So picking horizon to be greater than $n=\frac{1}{\sigma_0^2} = 2500$, we can see both scenarios. 

Optimal action selection rate plots shown in Figure \ref{fig:exp_all_share_vs_no_action} can additionally provide the same conclusion as discussed in Section \ref{sec:experiment}.




\end{document}